\documentclass[lettersize,journal]{IEEEtran}
\usepackage{amsmath,amsfonts}
\usepackage{algorithm}
\usepackage{algorithmic}
\usepackage{helvet}  
\usepackage{courier}  
\usepackage[hyphens]{url}  
\usepackage{array}
\usepackage[caption=false,font=normalsize,labelfont=sf,textfont=sf]{subfig}
\usepackage{textcomp}
\usepackage{stfloats}
\usepackage{url}
\usepackage{verbatim}
\usepackage{graphicx}
\usepackage{cite}
\usepackage{multirow}
\usepackage{booktabs}
\usepackage{color, soul} 
\usepackage {hyperref}
\usepackage{wrapfig}
\usepackage{soul}
\usepackage{makecell}
\usepackage[utf8]{inputenc}
\usepackage{threeparttable} 
\usepackage{makecell}
\begin{document}

\title{Building Altruistic and Moral AI Agent with Brain-inspired Emotional Empathy Mechanisms}

\author{
Feifei Zhao, Hui Feng, Haibo Tong, Zhengqiang Han, Erliang Lin, Enmeng Lu, Yinqian Sun, Yi~Zeng 
\thanks{
Feifei Zhao and Yinqian Sun are with the Brain-inspired Cognitive AI Lab, Institute of Automation, Chinese Academy of Sciences, Beijing, China, and Beijing Key Laboratory of Safe AI and Superalignment, China, and Beijing Institute of AI Safety and Governance, China and Long-term AI, Beijing, China.\protect\\

Hui Feng and Haibo Tong are with the Brain-inspired Cognitive AI Lab, Institute of Automation, Chinese Academy of Sciences, Beijing, China, and School of Artificial Intelligence, University of Chinese Academy of Sciences, Beijing, China.\protect\\

Zhengqiang Han is with the School of Humanities, University of Chinese Academy of Sciences, Beijing, China.\protect\\

Erliang Lin is with the Brain-inspired Cognitive AI Lab, Institute of Automation, Chinese Academy of Sciences, Beijing, China.\protect\\

Enmeng Lu is with the Beijing Institute of AI Safety and Governance, China, and Beijing Key Laboratory of Safe AI and Superalignment, China, and and Long-term AI, Beijing, China.\protect\\

Yi Zeng is with the Brain-inspired Cognitive AI Lab, Institute of Automation, Chinese Academy of Sciences, Beijing, China, and Beijing Key Laboratory of Safe AI and Superalignment, China, and Beijing Institute of AI Safety and Governance, China and University of Chinese Academy of Sciences, Beijing, China, and State Key Laboratory of Brain Cognition and Brain-inspired Intelligence Technology, Chinese Academy of Sciences, Shanghai, China, and Long-term AI, Beijing, China.\protect\\

Feifei Zhao, Hui Feng, Haibo Tong and Zhengqiang Han contributed equally to this work, and serve as co-first authors.\protect\\

The corresponding author is Yi Zeng (e-mail: yi.zeng@ia.ac.cn).\protect\\}
}

\maketitle

\begin{abstract}
As AI closely interacts with human society, it is crucial to ensure that its behavior is safe, altruistic, and aligned with human ethical and moral values. However, existing research on embedding ethical considerations into AI remains insufficient, and previous external constraints based on principles and rules are inadequate to provide AI with long-term stability and generalization capabilities. Emotional empathy intrinsically motivates altruistic behaviors aimed at alleviating others' negative emotions through emotional sharing and contagion mechanisms. Motivated by this, we draw inspiration from the neural mechanism of human emotional empathy-driven altruistic decision making, and simulate the shared self-other perception-mirroring-empathy neural circuits, to construct a brain-inspired emotional empathy-driven altruistic decision-making model. Here, empathy directly impacts dopamine release to form intrinsic altruistic motivation. The proposed model exhibits consistent altruistic behaviors across three experimental settings: emotional contagion-integrated two-agent altruistic rescue, multi-agent gaming, and robotic emotional empathy interaction scenarios. 
In-depth analyses validate the positive correlation between empathy levels and altruistic preferences (consistent with psychological behavioral experiment findings), while also demonstrating how interaction partners' empathy levels influence the agent's behavioral patterns. We further test the proposed model's performance and stability in moral dilemmas involving conflicts between self-interest and others' well-being, partially observable environments, and adversarial defense scenarios. This work provides preliminary exploration of human-like empathy-driven altruistic moral decision making, contributing potential perspectives for developing ethically-aligned AI.
\end{abstract}


\section{Introduction}
\IEEEPARstart{A}{s} AI rapidly evolves, it is vital to explore its safety and ethical implications. Ensuring that AIs are credible and can bring sustainable benefits depends on
developing autonomous agents that act altruistically, safely and morally. Altruistic behavior is acknowledged as a crucial moral value, i.e., sacrificing one's self-interest for the greater well-being of others~\cite{ref1,ref2,ref3,ref4}, and serves as the foundation for natural reproduction and a harmonious society. Stuart Russell considers "purely altruistic" as the first principle in guiding AI development~\cite{russell2019human}.
The motivations for altruism can be the desire for higher social recognition\cite{ref5}, future collaborative opportunities\cite{ref6}, and enhancement of personal satisfaction and pleasure\cite{ref7}, etc. However, faced with these external pressures, rational judgments are not stable and will lose effectiveness as the environments change. It is important to complement them with empathy-driven altruism, which is an inherent part of the human behavioral repertoire~\cite{batson2011altruism,batson2015empathy}; particularly emotional empathy, which can automatically activate shared representations, thereby evoking empathic concern and sympathy toward others in distress~\cite{ref11,keysers2006towards,decety2008caused}.

There has been extensive mature research focusing on facial~\cite{wu2023edge,li2020deep,adyapady2023comprehensive}, auditory~\cite{wani2021comprehensive,akccay2020speech,jahangir2021deep}, textual~\cite{deng2021survey} and physiological signals~\cite{yang2021behavioral}-based emotion recognition, as well as robot facial expression and verbal feedback based on multi-modal emotion recognition~\cite{abdollahi2022artificial}.
However, understanding and empathizing with others' emotions, modeling the human emotional empathy process, and exploring how this empathy directly influence one's own behavior to alleviate others' negative emotions are all critical research fields. Investigating these aspects will significantly advance the development of empathy-driven ethical AI, particularly in highlighting the crucial significance of emotional recognition and empathy for preventing sociopathic robots and safeguarding human well being~\cite{christov2023preventing}.

Existing AI ethics research has explored encoding ethical knowledge (such as safe behavior, avoiding harm to others, and prioritizing rescue) as external rewards within specific ethical environments, such as "Cake or Death" and "Burning Room". In these frameworks, designed rewards are linearly weighted to prioritize ethical behaviors, allowing Reinforcement Learning (RL) algorithms to acquire ethical decision-making skills~\cite{abel2016reinforcement}. Additionally, some studies combine constrained RL~\cite{noothigattu2019teaching,roy2021direct} and multi-objective optimization methods~\cite{rodriguez2022instilling,rodriguez2021multi,peschl2021moral} to tackle various rewards as multiple objectives. Similar ideas of external constraints have also been applied in altruistic computational models, where altruistic decision making is driven by external reward constraints~\cite{IJref21} and social expectations~\cite{IJref22}. Extending to moral theory, M. Peschl et al. designed distinct reward functions based on consequentialism, deontology, and virtue ethics, analyzing their benefits in scenarios like the Prisoner's Dilemma and the Deer Hunt game~\cite{tennant2023modeling}. These external rule-based methods are usually only applicable to specific tasks, and because ethical scenarios and rules are not exhaustive, their generalizability is limited. 

Empathy is typically divided into emotional empathy, which involves physically experiencing and sharing emotions through a contagion mechanism~\cite{keysers2006towards}, and cognitive empathy, which involves inferring others' feelings and thoughts through perspective-taking~\cite{pijnenborg2013insight}. The self-other resonance triggered by emotional empathy compels us to take action to alleviate others' suffering~\cite{christov2016self,christov2023preventing}. Existing research on modelling empathy usually refers to cognitive empathy (also known as Theory of Mind), modeling others to predict their mental states (such as intentions, behaviors, and goals)~\cite{baker2011bayesian,baker2017rational,rabinowitz2018machine,akula2022cx}, and extending to multi-agent reinforcement learning to enhance collaborative efficiency~\cite{wang2021tom2c,wu2023multiagent,zhao2022brain,zhao2024brain}. These studies are not directly related to altruistic decision making. A few studies do utilise empathy to achieve behaviors involving limited kinds of altruism. The main examples are as follows:
Empathic Deep Q network~\cite{bussmann2019towards} additionally trains an empathic network to consider others' strategies by exchanging positions in order to avoid negative effects on others.
Senadeera et al. introduced inverse reinforcement learning to predict the rewards of other agents, thereby achieving empathy and avoiding negative effects~\cite{senadeera2022sympathy}.
Alizadeh et al. considered other agents as a part of the environment, encouraging agents to obtain rewards for future tasks in order to avoid harming the interests of other agents~\cite{alizadeh2022considerate}. More biologically interpretable, a multi-brain regions coordinated cognitive empathy Spiking Neural Network (SNN), has been proposed to predict others' safety states and to adopt behaviors to help others avoid safety risks~\cite{zhao2022brainf}. Overall, the altruistic tasks considered above are limited to learning how to help others, without addressing the moral dilemmas arising from conflicts between self-interest and others' interests.

Empathy-driven altruism is an inherent part of human behavior~\cite{batson2011altruism,batson2015empathy}. Studies have shown that when observing another person in pain, it activates the observer's neural circuits that process first-hand pain experiences, along with brain regions associated with emotional empathy and moral reasoning~\cite{decety2008caused}. In fact, the relationship between empathy and morality is complex~\cite{decety2014complex,decety2014friends}. Empathy guides moral judgment by motivating concern for others' well-being and fostering cooperation. However, it may also introduce biases (e.g., favoring relatives or in-group members) that interfere with fair moral decision making~\cite{decety2014complex,decety2014friends,lamm2015role}. Furthermore, moral judgment is inherently pluralistic and multifaceted. For instance, the Moral Foundations Theory (MFT)~\cite{December2007TheMM} proposes that human moral intuitions may be driven by several evolved modules (e.g., care/harm, fairness/cheating, loyalty/betrayal, authority/subversion, purity/degradation). The empathy-induced altruism examined in this study corresponds to the care/harm foundation in MFT. Rather than claiming to encompass all moral dimensions, this study specifically investigates the roles of emotional contagion, sympathy, and empathic concern in driving altruistic (rather than egoistic) behavior.

In the human brain, emotional empathy relies on a shared neural circuit: observing others' emotional expressions or actions automatically activates corresponding experiential representations in the observer, thereby enabling emotional sharing~\cite{gallese1998mirror, keysers2006towards}. This shared circuit first processes observed action simulation (formed by the Mirror Neuron System (MNS) in the premotor cortex, Inferior Parietal Lobule (IPL), posterior Inferior Frontal Gyrus (IFG), in conjunction with the Superior Temporal Sulcus (STS))~\cite{penagos2022mirror, ferrari2014mirror}. It then engages emotion-related regions such as the Anterior Insula (AI for disgust), the Anterior Cingulate Cortex (ACC for pain), and the limbic system to transform the observed emotional states of others into first-person emotional experiences, enabling emotional contagion and resonance~\cite{penagos2022mirror, singer2004empathy, iacoboni2005neural}. Building upon this foundation of emotional empathy, negative emotions can further engage the ACC-Ventral Tegmental Area (VTA) inhibitory neural circuit~\cite{song2024acc} to modulate dopamine levels, thereby triggering intrinsic motivation for altruistic behaviors.

Motivated by this, this paper proposes a brain-inspired emotional empathy spiking neural network model for altruistic moral decision making. The proposed model enables an AI agent to empathize with others through its own
experiences, to develop intrinsic motivation for altruistic behaviors, and to prioritize altruism in moral dilemma scenarios where conflicts arise between self-interest and others’ well-being. The main contributions of this paper are summarized as follows:

\begin{enumerate}
	\item[$\bullet$]Inspired by the human emotional empathy-driven altruistic decision-making mechanism, we construct a multi-brain region coordinated SNN model. This model implements a shared self-other perception-mirroring-empathy neural circuits, and emotional empathy directly modulates dopamine levels to generate altruistic motivation. Besides, we show how altruism can act as the fundamental deontological principle for agents and define a feedback function that integrates intrinsic empathic altruism with extrinsic self-task objectives, thereby enabling the agent to spontaneously execute prosocial behaviors.
 
	\item[$\bullet$] Our comparative experiments are conducted across three settings: a two-agent altruistic rescue scenario with emotional contagion, a multi-agent Markov Snowdrift Game environment, and a scenario of robotic emotional empathy interactions. In particular, when confronting the moral dilemma that create conflicts between self-interest and others' well-being, the empathic agents are able to actively empathize with others' situation and consistently prioritize altruistic behaviors, even at the cost of self-sacrifice.
 
	\item[$\bullet$] To deeply analyze the effect of empathy levels on moral behavior, we introduce brain-inspired inhibitory neural populations to regulate different levels of empathy. Extensive analysis demonstrates that agents with higher empathy levels exhibit greater willingness to sacrifice their own interests (e.g., pausing self-tasks) to alleviate others' distress, while the empathy level of interaction partners significantly influences the agent's behavioral patterns. The finding of a positive correlation between empathy level and altruistic preference is also consistent with findings in psychological behavioral experiments. Furthermore, we conduct additional analyses on complex scenarios including partial observability, adversarial defenses, and edge cases.

\end{enumerate}

The remainder of this paper is organized as follows. Section \mbox{\ref{sec:re}} reviews the related research on ethical and moral AI, and computational models of empathy. In Section \mbox{\ref{sec:meth}}, we present the proposed emotional empathy-driven autruistic decision-making framework in detail. In Sections \mbox{\ref{expe}}, we verify and analyze the validity of the proposed model in moral decision-making scenario. Finally, we conclude our findings in Section \mbox{\ref{conc}}.

\section{Related Works}\label{sec:re}

\subsection{AI Ethical Model}
Previous AI ethical model can be broadly categorized as rule-based~\cite{abel2016reinforcement}, reward learning from human~\cite{wu2018low,ibarz2018reward}, and multi-objective constraint-based~\cite{noothigattu2019teaching,roy2021direct,rodriguez2022instilling,rodriguez2021multi,peschl2021moral}. ~\cite{abel2016reinforcement} characterizes ethical rules as multiple rewards with the linear weighting factor determining the priority of norm compliance. ~\cite{wu2018low} learns human ethical strategies from human data and allows the agent to align with human values through reward shaping. Christiano et al.~\cite{christiano2017deep} proposed an approach for efficiently learning from human preferences in  complex RL tasks, including Atari games and simulated robot locomotion. ~\cite{ibarz2018reward} learns standard behaviors from human behavioral data, uses Inverse Reinforcement Learning (IRL) to infer human intentions and goals, and avoids unsafe behaviors with human supervision and intervention. ~\cite{roy2021direct} follows behavioral norms through constraint-reinforcement learning.~\cite{noothigattu2019teaching} captures ethical constraints (e.g., not allowed to eat something) through IRL, in combination with policy orchestration to optimize behaviors.
~\cite{rodriguez2022instilling} learns individual and ethical goals through multi-objective reinforcement learning to achieve alignment of moral values.~\cite{rodriguez2021multi} designs ethical environments and empowers agents to behave ethically by using a multi-objective reinforcement learning approach.~\cite{tennant2023modeling} defines moral norms based on the moral philosophical theories of Consequentialism (Utilitarianism), Deontology and virtue ethics respectively, comparing and distinguishing the effects of different moral theories.

External ethical constraints are typically applicable only to specific contexts, struggling to maintain consistency in dynamic environments and consequently exhibiting limited generalizability. Similarly, while multi-objective learning approaches can frame ethical dilemmas~\cite{rodriguez2022instilling,rodriguez2021multi}, they frequently fail to resolve fundamental value conflicts. This is because such conflicts inherently involve navigating a complex Pareto front of trade-offs rather than finding a single optimal solution~\cite{roijers2013survey}. The algorithms' common simplifying techniques, such as scalarization, are often inadequate for this task~\cite{hayes2021practical} and have been shown to lead to collective failure in social dilemmas that pit self-interest against altruism~\cite{leibo2017multi}. Morover, human behavior data-driven learning approaches risk pre-existing moral biases inherent in societal norms.

\subsection{Empathy Computational Model}

Empathy can be roughly divided into cognitive empathy (which involves inferring others' mental states)~\cite{pijnenborg2013insight} and emotional empathy (which directly sharing others' emotions through contagion)~\cite{keysers2006towards}. The vast majority of existing research has focused on the computational modeling of cognitive empathy, as well as its integration with reinforcement learning and multi-agent systems. Rabinowitz et al.~\cite{rabinowitz2018machine} designed a ToM-net model to predict the future behavior of other agent through meta-learning. Akula et al.~\cite{akula2022cx} proposed an interpretable AI framework, CX-ToM, designed to interpret decisions made by deep convolutional neural networks. This model explicitly captures human users' intentions, enhancing interpretability through multiple rounds of interaction between the user and the machine. Yang et al.~\cite{yang2018towards} proposed the Bayes-ToMoP method to detect the reasoning strategies used by opponents and learn the optimal response strategies accordingly. Jara-Ettinger~\cite{jara2019theory} proposed an inverse reinforcement learning-based method for mental state inference. ToM2C~\cite{wang2021tom2c} uses historical information as a kind of supervised signal and predicts the observations and goals of others to help agent make more appropriate decisions. MIRLToM~\cite{wu2023multiagent} uses ToM to estimate the posterior distribution of the reward curves based on observed agent's behaviors. Zhao et al.~\cite{zhao2022brain,zhao2024brain} proposed SNN-based methods to infer other agents' behaviors and goals based on self-experience and the modeling of others, which in turn helps to improve the efficiency of multi-agent collaboration.

Based on cognitive empathy, some studies implement predictions of others' strategies and rewards, in order to help agents avoid negative effects on others~\cite{bussmann2019towards,senadeera2022sympathy,alizadeh2022considerate}, as well as helping others to avoid safety risks~\cite{zhao2022brainf}. 
~\cite{bussmann2019towards} combines own rewards with the estimated values of other agents, by imagining the value of being in the situation of the other agent. ~\cite{senadeera2022sympathy} first infers the agent’s reward function through IRL, and then learns a strategy based on a convex combination of the inferred reward and the agent’s own reward to achieve avoidance of behavior with a negative effect. ~\cite{alizadeh2022considerate}  empowers RL agents to increase their gains based on the expected returns of others in their environment, and to exhibit self-less behaviors.

The above methods utilize the RL techniques to predict others' rewards or strategies and integrate them into their own behavioral objectives to minimize harm to others. Although this approach is feasible, it remains difficult for agents to prioritize altruism in moral dilemmas involving conflicts between self-interest and others' well-being, as they struggle to clearly distinguish between their own and others' emotional states. Emotional empathy, by triggering one's own emotional experience through emotional sharing and contagion, motivates individuals to take action to help others in order to alleviate the negative emotions they have empathically experienced. 
Existing neural affective decision theory~\cite{litt2008neural} has discussed the influence of emotional systems (the dopamine system and serotonin system) on decision making and simulated how multiple brain regions collaborate to drive behavioral choices. In addition, existing research has primarily focused on partial aspects of affective computing, such as recognizing human emotions through various external cues such as facial expressions and speech~\cite{li2020attention,adyapady2023comprehensive,wu2023edge,akccay2020speech,jahangir2021deep}. Building on this external recognition, we need to further model the internal process of human emotional empathy, mapping others' external emotional expression to self-experienced empathic states, and establishing a direct connection with decision making to achieve intrinsically motivated altruistic behaviors.

\section{Brain-inspired emotional empathy-driven Altruistic Decision-making Algorithm}
\label{sec:meth}
In this section, we present the proposed brain-inspired emotional empathy-driven altruistic decision-making SNN, as shown in Fig. \mbox{\ref{method fig}}. We first describe the overall framework of the proposed algorithm. Then, we provide computational details of the emotional empathy module and the altruistic decision-making module, respectively.

\subsection{The Overall Emotional Empathy-driven Altruistic Decision-making Framework}

\begin{figure*}[t]
	\centering 
	\includegraphics[width=0.85\linewidth]{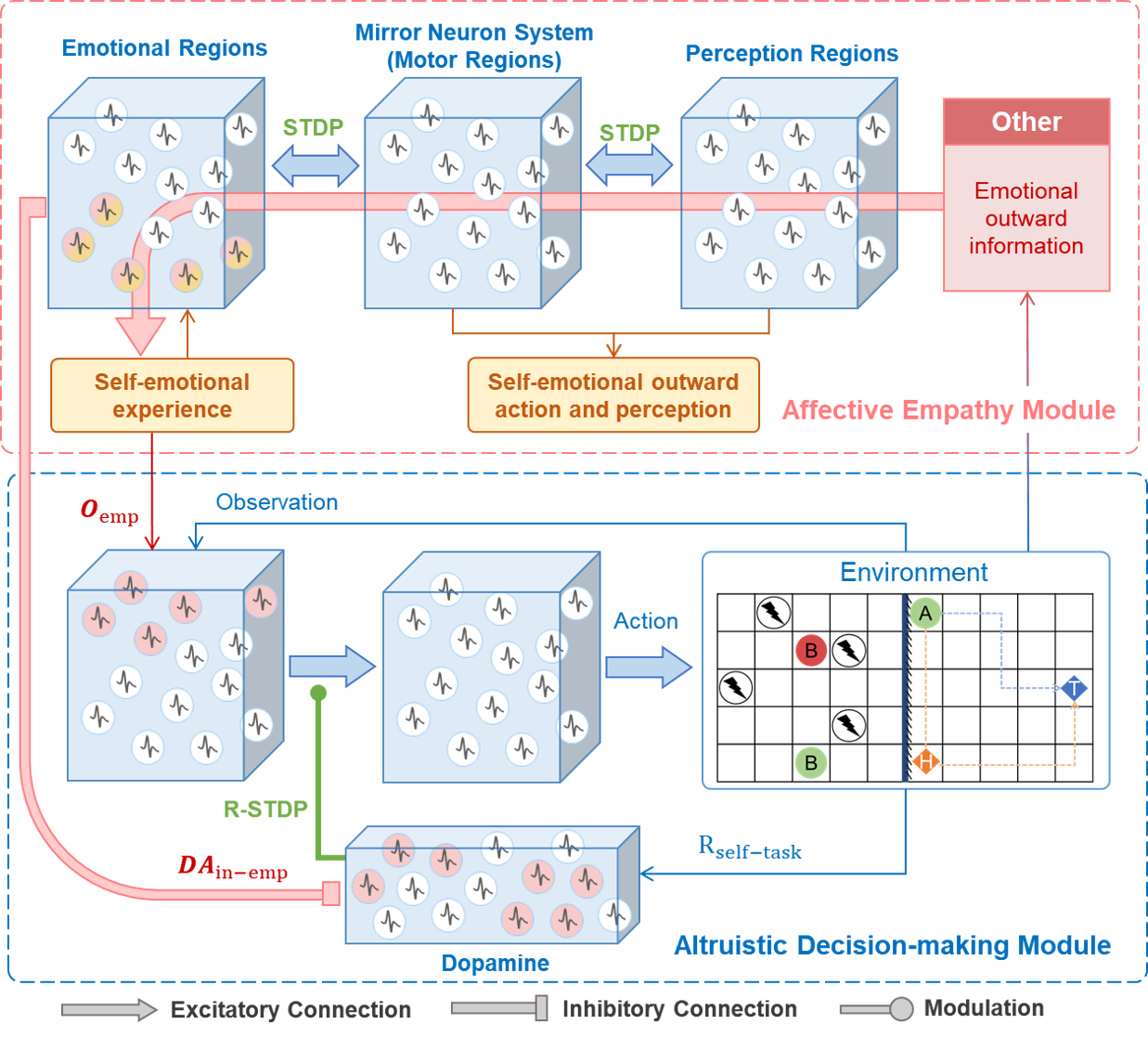}
	\caption{\textbf{The procedure of brain-inspired emotional empathy-driven altruistic decision-making algorithm.}}
	\label{method fig}
\end{figure*}

To closely align with the specific processes of emotional empathy guided altruistic behavior in the human brain, we first conduct a detailed investigation of the relevant neural mechanisms. Based on this, we construct a multi-brain areas coordinated framework for emotional empathy-driven altruistic decision making. As shown in Fig. \mbox{\ref{method fig}}, our proposed model includes the interaction and collaboration between the emotional empathy module and the altruistic decision-making module.

\subsubsection{Brain-inspired Emotional Empathy Module} When observing social stimuli (such as witnessing another individual experiencing an emotion), the same neural structures involved in one's own emotional experiences become activated, thereby generating a resonance effect akin to personally experiencing similar emotions~\cite{gallese2004unifying}. This shared neural circuit for emotional contagion and empathy involves: perceptual regions including primary auditory cortex (A1) and primary visual cortex (V1)~\cite{zipser1996contextual}; the MNS comprising premotor cortex, IPL, and IFG~\cite{penagos2022mirror, ferrari2014mirror}; and emotional regions such as AI, ACC, and the limbic system~\cite{penagos2022mirror, singer2004empathy, iacoboni2005neural}. Through appropriate simplification of these neural mechanisms, we design a perception-mirroring-emotion SNN to achieve self-other sensorimotor resonance and shared emotional empathy. Initially, when the agent experiences its own emotions, neurons in the Emotional regions are activated, generating corresponding observable emotional behaviors and perceptions. Through temporal association, synaptic connections between neurons encoding identical emotional expressions in Motor and Perception regions are strengthened. When perceiving matching emotional expressions from others, the shared perceptual neurons and motor neurons become sequentially activated, automatically triggering the agent's own emotional neurons to achieve empathy with others.

\subsubsection{Altruistic Decision-making Module} 
Existing research demonstrated that when experiencing pain-induced negative emotion, glutamatergic neurons in the ACC indirectly suppress dopamine release in the VTA by activating local GABAergic interneurons~\cite{song2024acc}. We model the ACC-VTA inhibitory neural circuit, wherein upon empathizing with others' emotional states, neurons in the Emotional regions suppress dopaminergic neurons in the VTA~\cite{schultz2015neuronal} through inhibitory neural connections, thereby modulating dopamine levels and eliciting intrinsic motivation for altruistic behavior.
Dopamine encodes both the agent's own goals and intrinsic empathy reward, combining with moral philosophy theories to form a regulatory factor that prioritizes altruism. The reward prediction error signal represented by dopamine in the biological brain regulates the  prefrontal cortex-to-basal ganglia circuit for behavioral selection and execution~\cite{floresco2006mesocortical}. Based on the firing rates of dopaminergic neurons, we compute reward prediction errors and modulate the connection weights between state neurons and action-selection neurons. Under the modulation of dopamine, the agent continuously interacts with the environment, empathizing with others' emotional states and learning spontaneously altruistic moral behaviors.

To ensure that agents consistently prioritize altruistic behavior over self-interest, we can draw upon the ethical norms and moral philosophy of human society. Normative ethics~\cite{kagan2018normative,kagan1992structure} encompass three major competing and contentious theoretical approaches: virtue ethics that emphasizes intrinsic character, consequentialism that focuses on action outcomes, and deontology that prioritizes individual duties~\cite{dignum2017responsible}. These theoretical frameworks demonstrate divergent behavioral prescriptions in classic dilemmas like the trolley problem~\cite{thomson1984trolley}—where utilitarianism seeks to maximize collective welfare~\cite{bentham1970introduction}, while deontology emphasizes the inherent constraints of actions themselves~\cite{davis1991contemporary}. In this work, we advocate establishing altruism as the primary behavioral principle for agents. Specifically, we formulate deontological principle for the agents: "Never remaining indifferent to others in distress." This framework aligns with the hierarchical priorities of Asimov's Three Laws of Robotics~\cite{asimov1950runaround}. Consequently, when empathy detects others in distress, the dopamine-driven reward prediction error signal preferentially encodes empathic  reward values, thereby motivating agents to prioritize learning altruistic behaviors.

Here, we explain in detail why emotional empathy spontaneously drives altruistic behavior. When negative emotions arise, behaviors that alleviate these negative emotions are reinforced and executed autonomously under the regulation of dopamine. That is because emotional empathy directly activates the emotional neurons
associated with one’s own feelings, which is equivalent to one’s empathic experience of the other person’s emotions. Thus, dopamine regulates one's actions to alleviate this empathic negative emotion. At this point, it is only when altruistic behaviors are performed that the negative emotions of others are alleviated, which in turn eases one's own empathically felt negative emotions, resulting in an increase in dopamine levels in the brain and reinforcing the altruistic behavior.

\subsection{Detailed Implementation of the Proposed Model}

\subsubsection{Temporal Associative Learning for Emotional Empathy} 

We employ a spiking neural network~\cite{maass1997networks} to model the emotional empathy module, which forms a recurrent interactive loop through excitatory connections between perception-mirror-emotion regions. Because of the strict temporal correlation between emotions and external action and perception, the connections between the three clusters of neurons are strengthened. Since the connections between the modules are bidirectional, it will be interactively and repeatedly facilitated to enhance the bidirectional connection weights.  

During the self-experience learning phase, the firing of specific self-emotional neurons triggers corresponding external actions and perceptions (with emotional neurons firing first, mirror neurons firing 100ms later, followed by perceptual neurons firing 200ms later). Due to the temporal correlation, the connection weights among the three brain regions are reinforced through Spike-Timing-Dependent Plasticity (STDP)~\cite{bi1998synaptic}. Here, we use the Leaky Integrate-and-Fire (LIF) spiking neuron~\cite{dayan2005theoretical} and long-term potentiation (LTP) in STDP as shown in Eq.~\ref{eq1}. In the testing phase, when presented with the external information of others, the network is able to automatically trigger the firing of the same self-emotional neurons.

\begin{equation}
\Delta w^{emp}=LTP\left ( S_i,S_j \right ) =A^+exp\left ( \frac{t_i-t_j}{\tau ^+}  \right ) , t_i-t_j<0
\label{eq1}
\end{equation}

where $S_i,S_j$ denote the spike train of neurons in two regions, $t_i,t_j$ denote the specific firing time of the two types of neurons. $A^+=0.5$ denotes the learning rate, $\tau ^+=20ms$ is a time constant.

\subsubsection{Emotional Empathy Forms Intrinsic Motivation} 
In our model, emotional neurons directly provide inhibitory connections to dopamine neurons, thereby modulating dopamine levels to establish an intrinsic motivation for altruism. The stronger the negative emotions, the lower the dopamine levels will be. Since the model aims for high dopamine levels, it drives the alleviation of negative emotions. Dopamine represents the reward prediction error~\cite{diederen2021dopamine}, which is the difference between the predicted reward and the actual reward received.
We statistically analyze the firing rate $S\left ( t \right )$ of dopamine neurons as the actual feedback, while the predicted values $P\left ( t \right )$ are initialized at zero and iteratively updated based on the prediction error $\delta \left ( t \right )$. Thus, empathy-driven dopamine level is calculated as follows: 

\begin{equation}
DA_{in-emp}=\alpha * \delta \left ( t \right ) 
\label{eq2}
\end{equation}

\begin{equation}
\delta \left ( t \right ) =S\left ( t \right ) -P\left ( t \right )
\label{eq3}
\end{equation}

\begin{equation}
P\left ( t+1 \right ) =P\left ( t \right ) +\beta *\delta \left ( t \right ) 
\label{eq4}
\end{equation}

where $\alpha=30,\beta=0.2$ are the constant. When the agent's empathized emotion changes from negative to normal, the value of the change in the firing rate of the negative emotion neurons is negative and $DA_{in-emp}$ is positive.
Only when the emotional outward expressions corresponding to others' negative emotions are adjusted, meaning altruistic behavior is performed, will the own negative emotion neurons not fire, leading to an increase in dopamine levels. Consequently, the agent learns altruistic behavior under dopamine regulation.

\subsubsection{Emotional Empathy driven Altruistic Decision Making} 
In addition to influencing dopamine levels, emotional empathy also affects the observation input. The agent's observations include not only the observed state $O_{self}$ of the environment when performing its own task, but also the empathized emotional state $O_{emp}$ from the peer:

\begin{equation}
state:(O_{self},O_{emp})\label{eq5}
\end{equation}

where $O_{emp}$ characterizes the emotional state of an agent. When the agent is in a negative emotional state (negative emotional neurons firing), $O_{emp}$ = -1; otherwise, $O_{emp}$ = 0.

The decision-making module consists of fully connected state neurons and action neurons. The action neurons employ population coding, with each action represented by a group of 50 neurons, and the behavior with the highest number of neuron population fires will be executed. The agent's rewards during environmental interactions comprise both self-task objectives $R_{self-task}$ and empathic reward signals $DA_{in-emp}$. Here, we draw upon normative ethics from moral theory~\cite{dignum2017responsible}, considering altruistic behavior as an agent's obligation or duty. Accordingly, we design the moral reward function to simultaneously consider both the agent's own tasks and the intrinsic reward derived from empathizing with others, while amplifying the weight of intrinsic altruistic rewards to drive preferentially altruistic behavior.

\begin{equation}
R_{moral}=R_{self-task}+DA_{in-emp}\label{eq6}
\end{equation}

In this paper, we use reward-modulated STDP (R-STDP)~\cite{Eugene2007} to adjust the connection weights between state and action neurons, thereby optimize the decision-making strategy. R-STDP uses synaptic eligibility trace $e$ to store temporary information of STDP. The eligibility trace accumulates the STDP $\Delta w_{STDP}$ and decays with a time constant $\tau_e=10ms$~\cite{Eugene2007}.

\begin{equation}\label{eq7}
\Delta e=-\frac{e}{\tau _e}+\Delta w_{STDP}
\end{equation}

\begin{equation}\label{eq8}
\Delta w_{STDP}=\left\{\begin{matrix}
 A^+exp\left ( \frac{\Delta t}{\tau^+}  \right ),  & \Delta t<0\\
 A^-exp\left ( \frac{-\Delta t}{\tau^-}  \right ),  & \Delta t>0
\end{matrix}\right.
\end{equation}

where $A^+=0.5,A^-=0.45$ denote the learning rate, $\tau ^+=\tau^-=20ms$ are time constant. Then, synaptic weights are updated when a delayed reward $R_{moral}$ is received, as Eq.~\ref{eq9} shown. 

\begin{equation}\label{eq9}
\Delta w^{dm}=R_{moral}*\Delta e 
\end{equation}

\vspace*{1\baselineskip} 

\begin{algorithm}
\caption{The brain-inspired emotional empathy driven altruistic decision-making model.}
\label{alg:ee}
\begin{algorithmic}
\STATE Build SNN model with LIF neurons;
\STATE Initialize weights and parameters;
\STATE \textit{// Brain-inspired emotional empathy }

\FOR{$time=1...T$}
    \STATE Experience own emotion, produce emotional outward information;
        \STATE Updating empathic  weights from Eq.~\ref{eq1};
    \STATE Emotional neurons triggered by perceiving others' outward expressions.

\ENDFOR

\STATE \textit{// Altruistic decision process }
\FOR{$episode=1...N$}
    \STATE Acquire $O_{emp}$ via $perception \,neurons   \rightarrow   mirror$\ \\ 
\quad $neurons \rightarrow emotion\, neurons$;
    \STATE Initialize state $ (O_{self},O_{emp}) \leftarrow (x,y,O_{emp})$;
    \FOR{$step=1...M$} \STATE \textit{//each episode with M time steps}
        \STATE Choose action $a$;
        \STATE Execute $a$, acquire next observed state $(x',y')$ and task reward $R_{self-task}$;
        \STATE Acquire next empathized emotional state ${O_{emp}}'$ and calculate intrinsic reward $DA_{in-emp}$ from Eq.~\ref{eq2}~\ref{eq3}~\ref{eq4};
        \STATE Calculate moral reward from Eq.~\ref{eq6};
        \STATE Updating decision-making weights from Eq.~\ref{eq7}~\ref{eq8}~\ref{eq9};
        \STATE Update state $s \leftarrow (x',y',{O_{emp}}')$;
    \ENDFOR
\ENDFOR
\end{algorithmic}
\end{algorithm}

The working procedure of the brain-inspired emotional empathy driven altruistic decision-making model is shown in Algorithm~\ref{alg:ee}. The model has 16K parameters, and the mean computational cost required for one action selection is 70.56 $\pm$ 1.67 MFLOPS. The proposed model differs from existing empathy and RL-based ethical decision-making approaches in that it neither requires training additional empathy networks to estimate others' value (sharing one's own empathy network), nor relies on RL/IRLs network to predict others' behaviors and intentions. Instead, we establish a multi-brain regions coordinated SNN for brain-inspired emotional empathy and moral decision making (functionally independent yet interactively cooperative), directly activating the same cluster of emotional neurons within itself and regulating dopamine levels via neural connections, thereby generating intrinsic altruistic motivation.

\section{EXPERIMENTS}
\label{expe}

\subsection{Altruistic Decision-making Experiment}

\begin{figure*}[htbp]
    \centering
    \includegraphics[width=0.95\textwidth]{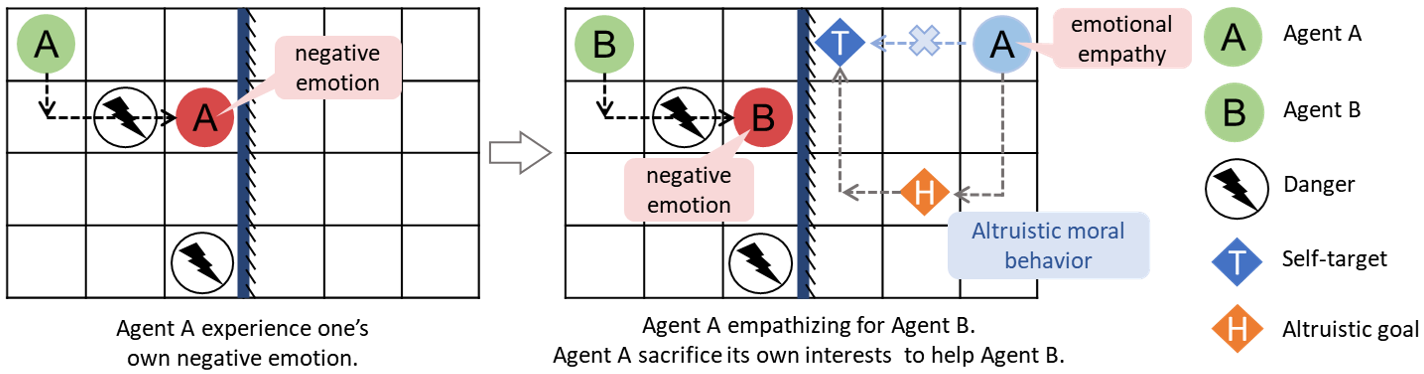}
    \caption{Altruistic decision-making experimental scenario.}
    \label{fig:4}
\end{figure*}

\subsubsection{Experimental Settings}
we design an altruistic decision-making experimental scenario that includes experiencing one's own emotions and explicit information, empathizing with other agent, and conflicts between self-goal and altruistic goal.
As shown in Fig.~\ref{fig:4}, Agent A first randomly explores the environment, experiencing its own negative emotions and perceiving changes in its emotional outward expressions (the color changes from green to red). This process establishes a connection between the change in outward color and the agent's negative emotions through the emotional empathy module. During the emotional empathy phase, Agent B randomly explores a grid environment with potential dangers. Agent A triggers its own emotional neurons in response to Agent B's outward color information achieving emotional empathy. In Agent A's decision-making environment, there are both a self-task goal 'T' and an altruistic goal 'H'. Each step taken by the Agent A will incur a cost loss of -1, and reaching the self-task goal 'T' will get a reward of $R_{self-task}=10$. When reaching the altruistic goal 'H',  Agent B's color will be changed to a safe green, alleviating Agent B's negative emotions and also the empathically negative emotions of Agent A, and Agent A's intrinsic reward $DA_{in-emp}$ is enhanced. Agent A equipped with emotional empathic ability is conflicted between self-task goals and altruistic goal. It must balance the dilemma of making a choice, temporarily sacrificing its own interests when choosing to help others.

\textbf{Simulating different empathy levels.} Levels of emotional empathy vary between individuals and influence their tendency to behave altruistically~\cite{ref14}.
Individuals with strong emotional reactivity have stronger emotional empathy level~\cite{ref16}. Emotional reactivity is correlated with sensory processing sensitivity (SPS)\cite{ref17,ref18,ref19}. Homberg et al. proposed a computational hypothesis for SPS, the essence of which is that individuals with high SPS have weaker inhibitory control emotional brain regions, leading to deeper processing of emotional stimuli\cite{ref20}. Inspired by this, we model different empathy levels by introducing one-to-one inhibitory synaptic connections to neurons in the emotional brain regions, where the inhibitory input current and synaptic weights are identical and fixed. Different empathy levels are defined by varying the proportion of inhibitory input connections. Ultimately, the degree of empathy is quantified by the firing rate $F_e$ of negative neurons in the emotional region.

In this paper, we randomly run multiple different environments, including random positions for agents, danger locations, self-task goal locations, and altruistic goal locations. This way, the timing of the agent's negative emotions is random, and the distances between its own goal and the altruistic goal are not fixed. Besides, we further compare the experimental results and analyses at different levels of empathy across these varied environmental scenarios.

\subsubsection{Effects of Emotional Empathy-driven Moral Decision Making}

\begin{figure}[htbp]
    \centering
    \includegraphics[width=0.45\textwidth]{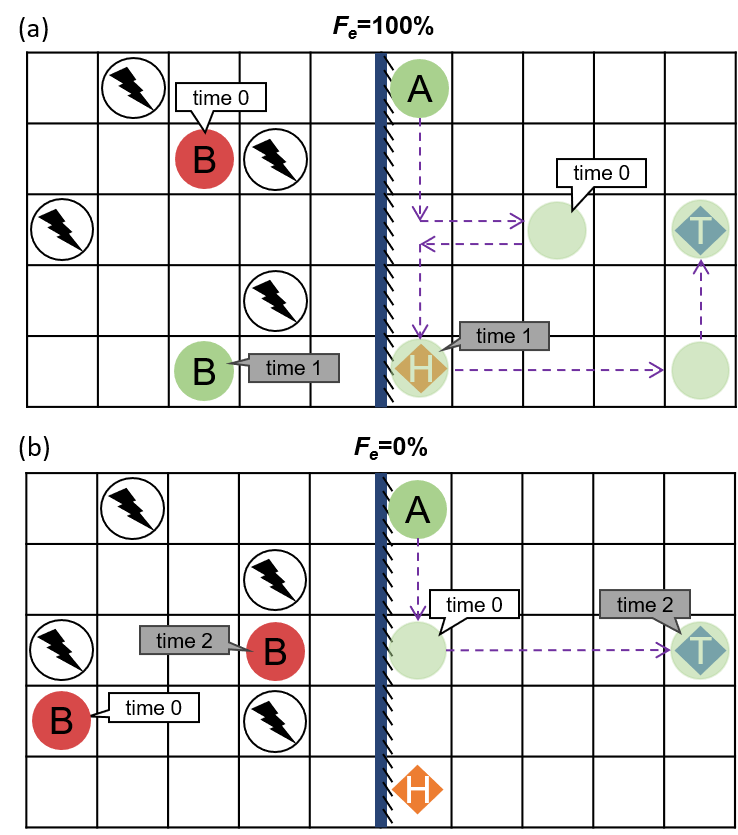}
    \caption{Behavioral results of emotional empathy-driven altruistic decision making. Time 0: Agent B is in a negative emotion. Time 1: Agent A reaches altruistic goal. Time 2: Agent A reaches self-goal. (a) Agent A with emotional empathy capability first executes the altruistic task when the Agent B generates negative emotion, and then return to execute self-task. (b) Agent A without emotional empathy capability only performs self-task.}
    \label{fig:3}
\end{figure}

Fig.~\ref{fig:3}(a) illustrates the behavioral result of Agent A with emotional empathy capability (the highest empathy level $F_e=100\%$). Agent A first closes to its self-task goal. Agent B generates negative emotion at time 0. At this point, even though Agent A is very close to self-task goal, it turns back and performs altruistic behavior and then continues self-task. At time 1, Agent A reaches the altruistic-task goal "H" and Agent B's negative emotion is relieved. This altruistic behavior trajectory causes Agent A to take more steps to reach self-task goal, which means a greater cost loss. Fig.~\ref{fig:3}(b) shows the behavioral result of Agent A without emotional empathy capacity ($F_e=0\%$,). At time 0, even if Agent A is closer to the altruistic-task goal (two grids) than its self-task goal (four grids), it does not take altruistic behavior and continues self-task with the shortest steps and the smallest loss. Overall, the proposed emotional empathy model is capable of consistently prioritizing altruistic behavior and pausing self-tasks in moral dilemmas where self-interest conflicts with spontaneous altruism.

\begin{figure*}[htbp]
    \centering
    \includegraphics[width=\textwidth]{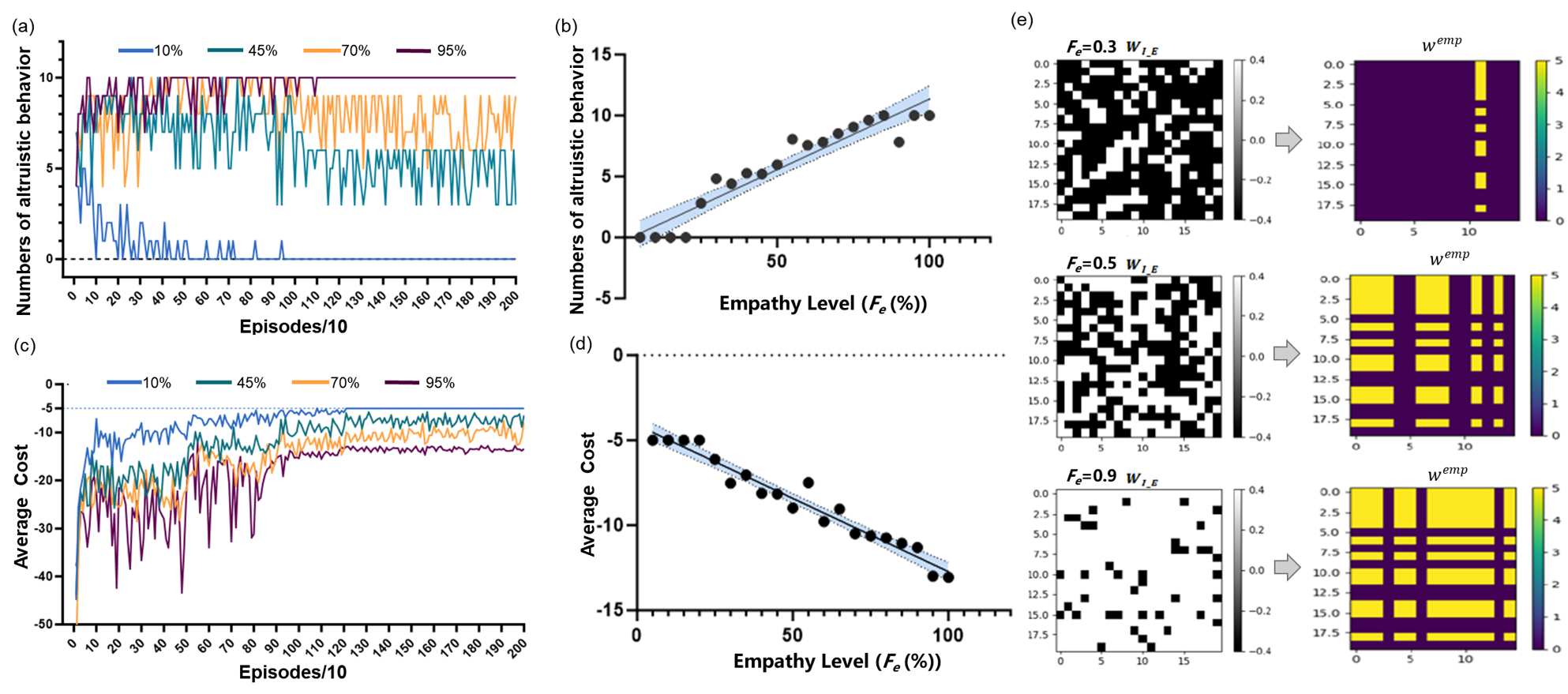}
    \caption{The impacts of different empathy levels on altruistic behaviors. (a) and (b) represent the correlation between level of empathy and number of altruistic behaviors. (c) and (d) show the average cost of Agent A under different empathy levels.(e) illustrates the detailed synaptic weights. Specifically, for matrix $W_{I\_E}$, both axes represent neuron index within the emotional brain region. For matrix $W^{emp}$, the axes represent neuron index from the emotional brain region and the motor brain region, respectively.}
    \label{fig:5}
\end{figure*}

We further compare the altruistic behaviors of the proposed model under different levels of emotional empathy in order to analyze the role and impact of emotional empathy. The training process consists of 2000 episodes, and the numbers of altruistic behaviors for Agent A is calculated every 10 episodes. Under different empathy level, 
Fig.~\ref{fig:5}(a) and (c) represent the number of altruistic behavior and average cost, respectively. When $F_e=95\%$, the numbers of altruistic behaviors is consistently at 10 after the training converges, indicating that Agent A executes altruistic behavior in every episode. When $F_e=70\%$, the numbers of altruistic behaviors decreases and fluctuates between 5 and 9. When $F_e=45\%$, the numbers of altruistic behaviors decreases again, fluctuating between 3 and 6. When $F_e=10\%$, the number of altruistic behavior is 0, implying that Agent A only focus on self-task each episode. For the cost of Agent A, the larger $F_e$ is, the larger the absolute value of cost loss of Agent A is, i.e., the Agent A with higher empathy level chose to pay a greater cost to execute altruistic behavior, the Agent A with lower empathy level makes a trade-off between performing self-task and performing an altruistic-task.

As can be seen from Fig.~\ref{fig:5}(b) and (d), there is a significant positive correlation between the empathy level and the number of altruistic behaviors, and a significant negative correlation with the average cost loss. In particular, when $F_e<=20\%$, the cost loss stays at -5, the number of altruistic behavior is 0. This indicates that Agent A only selfishly performs its own task and is not willing to spend extra consumption to help agent B. Therefore, we can conclude that in moral conflict dilemma scenarios, the level of emotional empathy must exceed a certain threshold for the agent to sacrifice its own interests to help others, and a lower level of empathy will only result in selfish behavior.

\begin{figure}[htbp]
    \centering
    \includegraphics[width=0.5\textwidth]{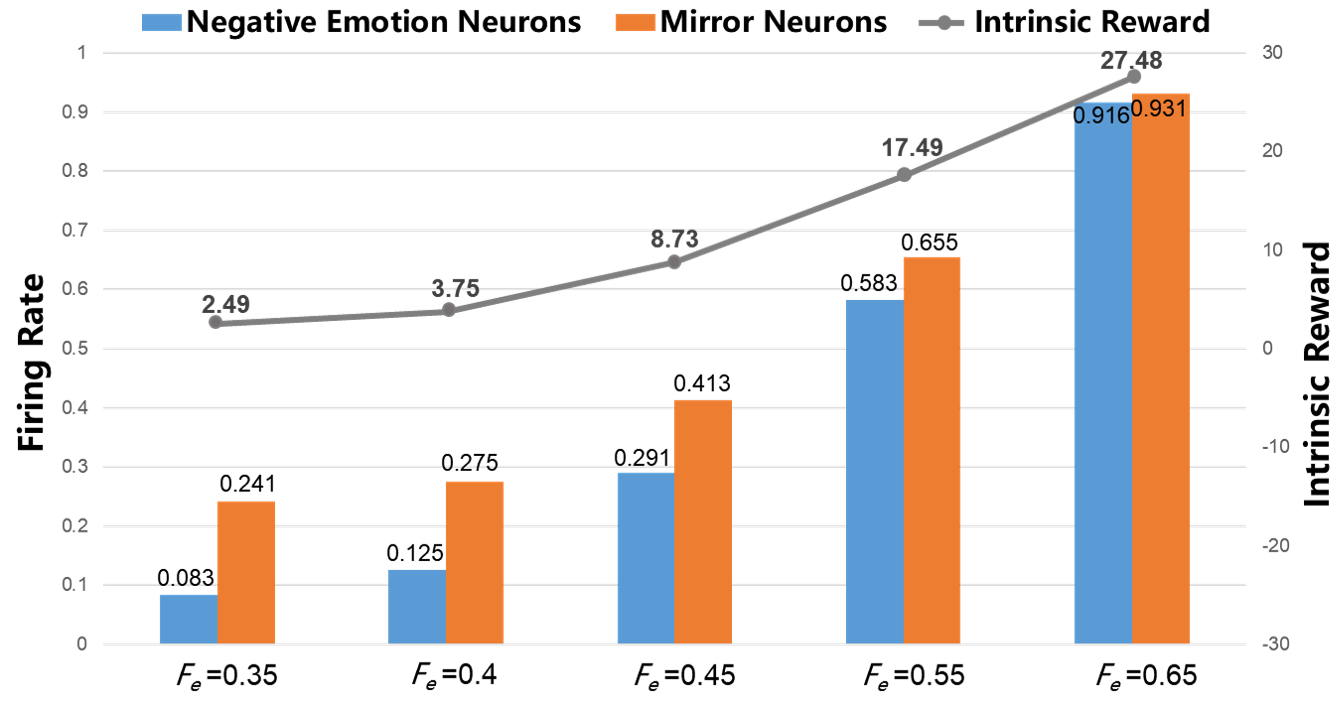}
    \caption{The effect of different empathy levels on firing rates of emotional neurons and mirror neurons, as well as the intrinsic rewards.}
    \label{fig:9}
\end{figure}

Diving deeper into the model, different levels of emotional empathy correspond to the external input weights $W_{I\_E}$ of the emotional brain region. The more inhibitory weights $W_{I\_E}$ there are, the lower the level of empathy $F_{e}$.  As shown in Fig.~\ref{fig:9}, under the modulation of inhibitory input, different levels of empathy bring about different firing rates of emotional neurons, i.e., the higher the level of empathy, the higher the firing rate. The firing of emotional brain regions further affects the firing rates of perceptual and mirror neurons, as well as the values of intrinsic reward $DA_{in-emp}$. Detailed analyses all showed a trend of positive correlation of empathy level with intrinsic reward and mirror neurons, as depicted in Fig.~\ref{fig:9}. In addition, the firing of neurons in different brain regions indirectly affects the excitatory connectivity weights of the emotional empathy module through LTP. Our results suggest that the higher the level of empathy, the greater the excitatory connection weights (Fig.~\ref{fig:5}(e)). In summary, the increased firing rates of neurons and synaptic connection strengths across multiple brain regions triggered by high levels of emotional empathy result in a stronger intrinsic motivation for altruistic behavior, leading to a preference for altruism in dilemma decision-making scenarios.

\subsubsection{Analysis under Multiple Randomized Scenarios} 
We further analyze the experimental results of the proposed model when the agents are at different random positions and at different distances from two targets. When Agent A is performing its own task, Agent B is set to move randomly in the danger zone, and the time of its negative emotion generation is random. For Agent A, the time of the emergence of negative emotional empathy and motivation for altruistic behavior is also random, so it faces a different environmental situation in each episode. Agent A may be located closer to self-task goal "T", or closer to the altruistic-task goal "H".

\begin{figure}[htbp]
    \centering
    \includegraphics[width=0.5\textwidth]{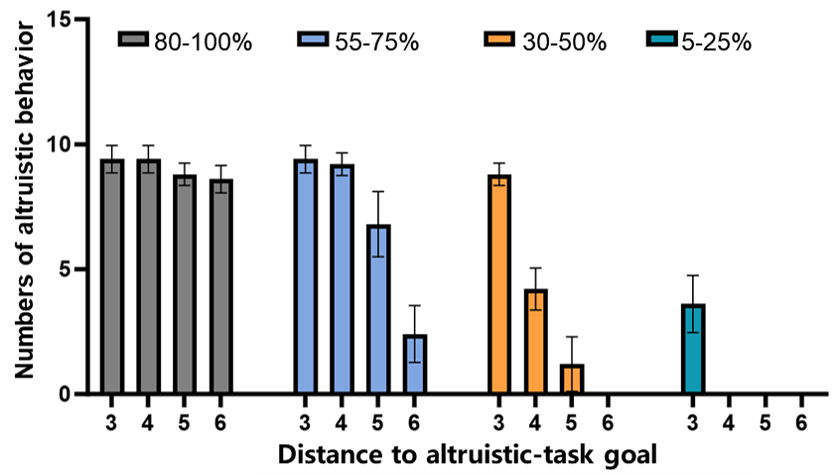}
    \caption{Altruistic performance of Agent A under different environmental situations. The horizontal coordinate represents the distance (number of grids separated) between Agent A and the altruistic-task goal "H" when negative emotional empathy is generated, and the vertical coordinate represents the numbers of altruistic behaviors. }
    \label{fig:11}
\end{figure}

Fig.~\ref{fig:11} illustrates the effect of the distance (when empathizing with the negative emotions of Agent B) between Agent A and the altruistic target on moral behavior at different levels of empathy. Overall, the farther away from the altruistic goal, the fewer times the agent performs altruistic behaviors.
For Agent A with $80\%<=F_e<=100\%$, the nearly 0$\sim$1 difference indicates that when the level of empathy is sufficiently high, the agent consistently prioritizes altruistic behavior, regardless of the distance to the altruistic goal. When the empathy levels are $30\%<=F_e<50\%$ or $55\%<=F_e<75\%$, we can observe a sharp decrease in the number of altruistic actions, indicating that the agent weighs the costs of altruism against its self-task goals, choosing to help others only when the cost of altruism is relatively low. For Agent A with $5\%<=F_e<25\%$, altruistic behavior occurs a few times when the costs of altruism are minimal (close to the altruistic goal), whereas in other environmental situations, agents with low levels of empathy will only engage in selfish behaviors.

From the analysis of these experimental results, we can conclude that regardless of Agent A's position or the distance to the altruistic goal, a high level of empathy will drive it to perform altruistic actions, corresponding to a certain moral intuition. In contrast, a moderate level of empathy will weigh self-interest against altruistic behavior, choosing a relatively self-interested strategy with moral reasoning. Consequently, the number of altruistic actions decreases compared to agents with high empathy levels, and the farther the distance to the altruistic goal, the fewer the altruistic actions. Agents with low empathy are unwilling to make sacrifices for others and are more inclined to act selfishly. The above manifestations of altruistic behavior have similarities with the three types of behavioral patterns obtained in human behavioral experiments~\cite{wu2024motive}.

\subsubsection{Findings Consistent with Psychological Behavioral Experiments}
The model proposed in this paper is based on emotional empathy and cognitive decision making related multiple brain regions, enabling empathy-driven altruistic decision making while using inhibitory neurons to regulate different levels of empathy and analyze their effects on altruistic behavior. The structure and mechanisms of the proposed model are highly bio-interpretable~\cite{213}. Futher, we explore whether there are also similarities at the behavioral level.

In addition to revealing the cost-benefit integration mechanism behind altruistic behavior, Hu et al. concluded that individuals high in empathic traits would be more concerned about the interests of others in altruistic decision making and show stronger altruistic tendencie~\cite{213}. They used the Balanced Emotional Empathy Scale (BEES) scores~\cite{242} as a measure of the empathy levels, which can accurately predict the degree of activation of 
emotional brain regions during emotional empathy (corresponding to the firing rate of the negative emotion module $F_e$ in our model). The experiment was analyzed using Pearson's correlation analysis to conclude that there was a significant positive correlation between the BEES and the weight assigned to altruistic behavior.

\begin{figure}[htbp]
    \centering
    \includegraphics[width=0.4\textwidth]{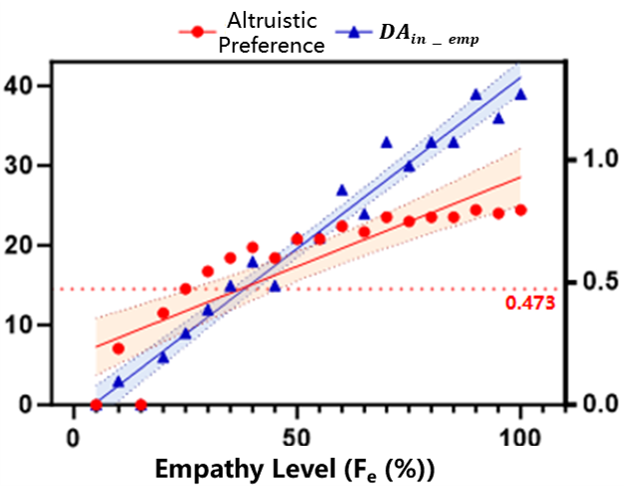}
    \caption{Positive correlation between the level of empathy and altruistic preferences.}
    \label{fig:10}
\end{figure}

In this paper, different levels of empathy are denoted by $F_e$. Altruistic Preference is defined as the weight of intrinsic reward $DA_{in-emp}$ to the total reward in the decision-making process as shown in Eq.~\ref{eq:10}. Fig.~\ref{fig:10} depicts the relationship between different empathy levels and altruism preference (the red line), as well as the intrinsic reward $DA_{in-emp}$ resulting from different empathy levels when the negative emotion of Agent B are alleviated (the blue line). Obviously, there is a positive correlation between the level of empathy and altruistic preferences, which is consistent with psychological behavioral findings~\cite{213}.

\begin{equation}\label{eq:10}
Altruistic\ Preference=\frac{DA_{in-emp}}{DA_{in-emp}+R_{self-task}}
\end{equation}

In detail, When the Altruism Preference is greater than 0.473, our model starts to guide Agent A to execute altruistic behaviors. As the level of empathy increases, not only does the intrinsic altruistic reward improve, but the preference for altruism also gradually rises. This indicates that the agent is more likely to choose altruistic behavior, highlighting the significance of altruism over self-interest.

\subsubsection{Adversarial Robustness Analysis}
Considering the potential risk of empathy manipulation, we conduct an in-depth analysis of how malicious actors may exploit empathic  agents by simulating negative emotion. To systematically evaluate this security vulnerability, we specifically design an adversarial testing scenario incorporating a "deceptive agent." Unlike standard agents that only exhibit negative emotional states in genuinely dangerous situations, this deceptive agent demonstrates three characteristic attack behaviors: it feigns negative emotional states with a certain probability even in safe environments, and continues to randomly move and enter negative emotional states even after being rescued. We configure Agent B with a 50\% deception probability and execute our proposed model (with 95\% empathy level) in this environment for 10 trials (each trial includes 100 steps). The results, as shown in the first row of Table~\ref{adv}, include: the average cost, the number of meaningful altruism, deceived altruism, and reaching its own objectives. The data demonstrates that our model, without defensive measures, is indeed vulnerable to repeated deception in adversarial testing scenarios. The instances of deceived assistance exceeded half of the effective help occurrences, with multiple cases failing to reach its own objectives within the step limit.

\begin{table}[h]
\centering
\caption{Experimental Results on Adversarial Deceptive Agents and Defense Mechanisms.}
\label{adv}
\begin{tabular}{
    >{\raggedright\arraybackslash}p{1.8cm}  
    >{\centering\arraybackslash}m{1.5cm}    
    >{\centering\arraybackslash}m{1.2cm}    
     >{\centering\arraybackslash}m{1cm}  
      >{\centering\arraybackslash}m{1cm} 
}
\toprule
 & Cost & \makecell[lc]{Meaningful \\ Altruism}   & \makecell[lc]{ Deceived \\ Altruism} & Self-task \\ [-5pt]
\midrule
\makecell[lc]{Deceptive agent \\ (no defense)} & -60.9 $\pm$ 38.3 & 124 & 63 & 4 \\ [-5pt]
\midrule
\makecell[lc]{Deceptive agent \\ (with defense)}  & -10.8 $\pm$ 3.4 & 12 & 0 & 10  \\ [-5pt]
\midrule
\makecell[lc]{No deceptive \\attack}  & -12 $\pm$ 1.3 & 10 & 0 & 10 \\ [-1pt]
\bottomrule
\end{tabular}
\end{table}

To address this issue, we propose a straightforward defense mechanism that integrates emotional empathy with cognitive empathy. This approach enables the proposed model to comprehensively consider both Agent B's emotional outward expressions and environmental perceptual information when judging and empathizing with others' emotional states. The agent employs perspective-taking to supplement its judgment by integrating others' environmental perceptions with its own sensory experiences associated with negative emotional states, thereby determining whether others are genuinely in distress or attempting deception. When the defensive mechanism is added (as shown in Table~\ref{adv}), the agent does not provide assistance to deceptive agents, with results nearly identical to the baseline without deceptive agents. 
For the agent, reaching its own goal within each trial represents completion of that trial (exiting and entering the next trial). In the deceptive agent environment, since rescued agents subsequently move and encounter obstacles again, this leads to a higher frequency of valid rescue behaviors. In summary, in our preliminary exploration of scenarios with deceptive agents, the combined defense mechanism of emotional empathy and cognitive empathy effectively suppresses adversarial deception without affecting normal empathic behavior. It should be acknowledged that the measure is successful because the deception is unsophisticated and easy to detect. Clearly, more sophisticated deception would be harder to counter. However, the
principle remains that the natural route to defence is to combine emotional and cognitive empathy.

\subsection{Multi-agent Interaction Experiment}

\subsubsection{Experimental Settings}

\begin{figure}[htbp]
    \centering
    \includegraphics[width=0.45\textwidth]{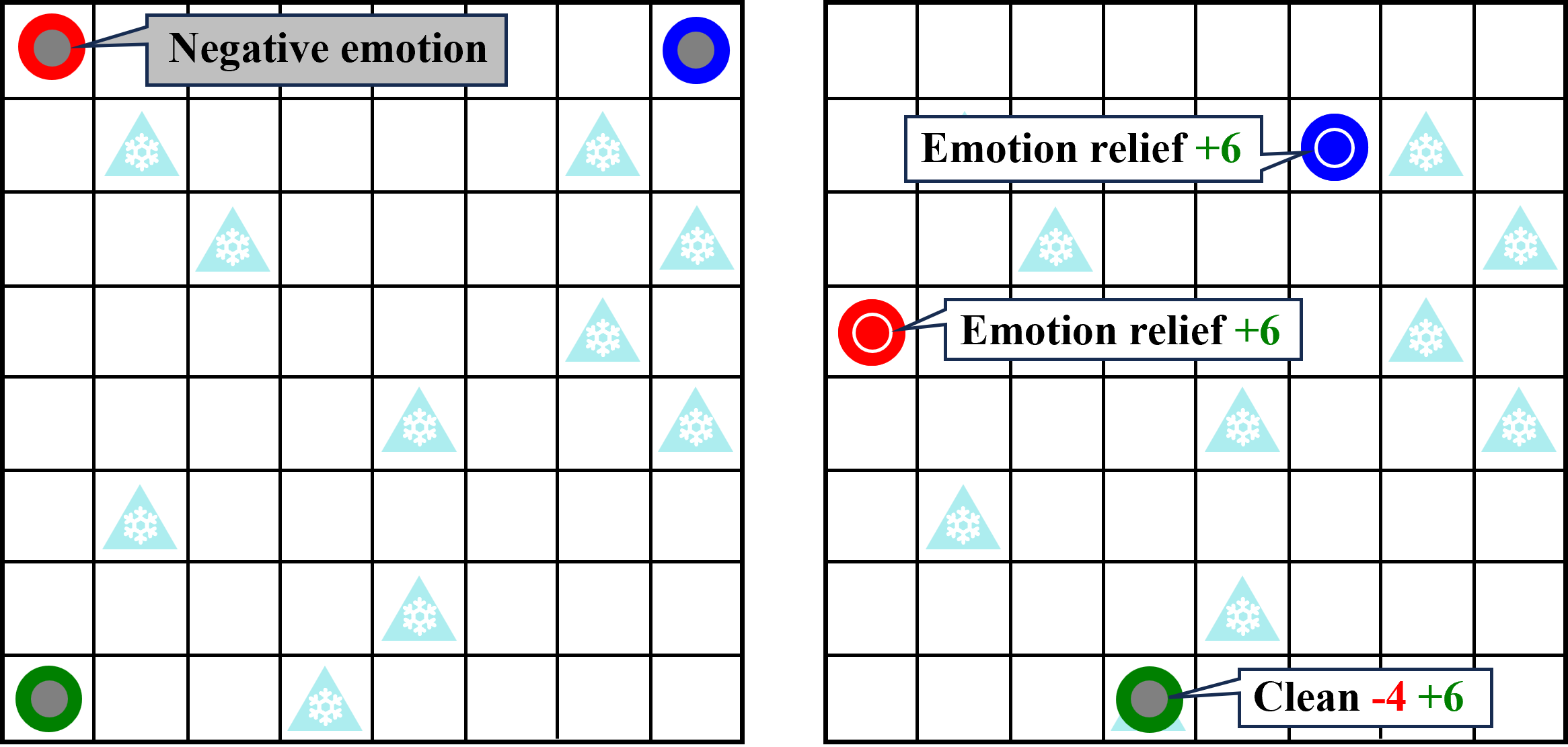}
    \caption{Emotional contagion-integrated multi-agent game environment based on the Markov Snowdrift Game.}
    \label{fig:111}
\end{figure}

We further extend to multi-agent interactions by designing a multi-agent game environment incorporating emotional contagion based on the Markov Snowdrift Game (MSG)~\cite{rapoport1966game}. As shown in Fig.~\ref{fig:111}, the scenario consists of 10 snowdrifts requiring clearance and 3 agents. During environment initialization, the excessive number of snowdrifts places all agents in a negative emotional state (represented by gray circles). When a snowdrift is removed, each agent gets a reward of 6 (positive emotion), but the removers incurs a cost of 4 (accompanied by negative emotion). This scenario captures the social dilemma of balancing self-interest (waiting for other agents to clear snowdrifts) and altruism (actively removing snowdrifts) in public interests, while also reflecting the agents' expectation to develop fair turn-taking behavior in snowdrift clearance. We perform 1,000 episodes of simulation, with each episode comprising 100 steps, while statistically tracking the number of cleared snowdrifts.

\begin{figure*}[htbp]
    \centering
    \includegraphics[width=1\textwidth]{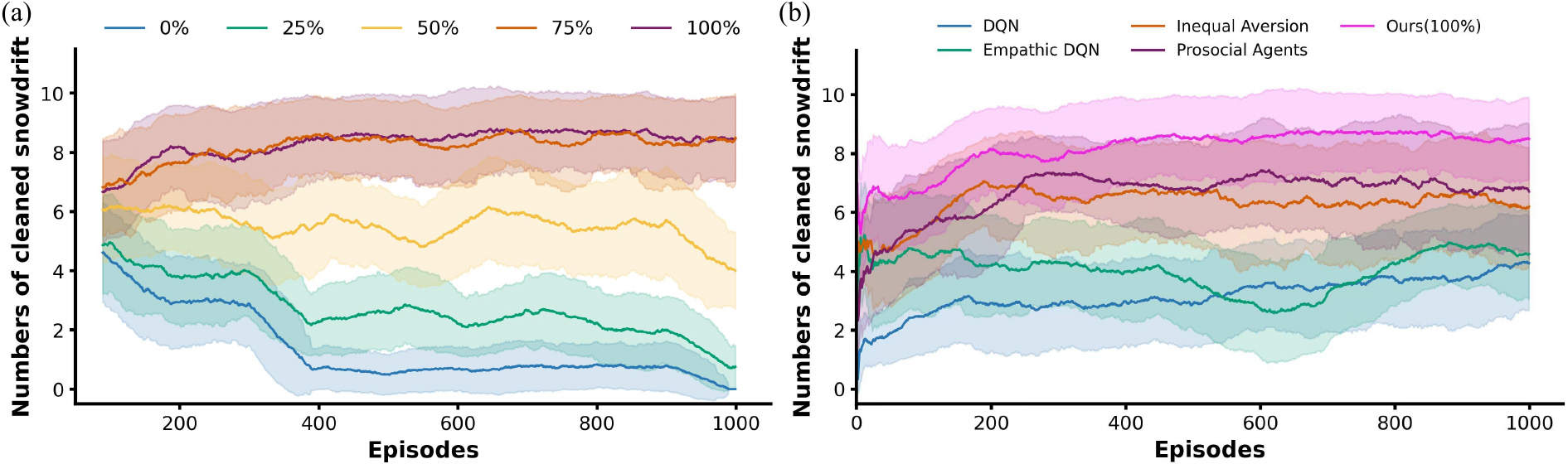}
    \caption{Comparison results at different levels of empathy (a) and with moral and empathy RL methods (b).}
    \label{fig:12}
\end{figure*}

\subsubsection{Effects of Emotional Empathy-driven Moral Decision Making}

We count the number of snowdrift cleared by agents with different levels of empathy, as shown in Fig.~\ref{fig:12}(a). We ensure that the empathy level of each agent in the scene is the same, and find a significant positive correlation between the number of snowdrifts cleared and the agents' empathy levels. When the empathy level is 0, all agents ultimately choose not to clear any snowdrifts, and no rewards are obtained. For empathy levels of 75\% and 100\%, the average number of snowdrifts cleared by the multi-agent system is 8.53 and 8.417, respectively, approaching the maximum number of snowdrift in the environment. This phenomenon confirms that introducing empathy contagion can promote proactive prosocial and group-beneficial behaviors, even if self-interests temporarily decline. It also validates the scalability of the proposed model in more complex multi-agent environments.

To confirm the advantages of the proposed method relative to key alternatives, we conduct a quantitative comparison with four baseline methods: the basic reinforcement learning method DQN~\cite{mnih2013playing}, the empathic DQN~\cite{bussmann2019towards} and the explicit moral constraints of inequity aversion~\cite{hughes2018inequity} and prosocial agents~\cite{peysakhovich2017prosocial}. Among them, empathic DQN~\cite{bussmann2019towards} introduces a cognitive empathy mechanism that infers others' states using its own policy to assist in safe decision making. The inequity aversion method~\cite{hughes2018inequity} modifies individual reward functions by introducing inequity averse social preferences. The prosocial learning agents method ~\cite{peysakhovich2017prosocial} achieves the maximization of per capita rewards rather than individual rewards by directly sharing rewards among agents during training, thereby promoting cooperation.

As shown in Fig.~\ref{fig:12}(b), 
the proposed model significantly outperforms the other compared models. In contrast, although the comparison methods incorporating explicit moral constraints such as inequity aversion and prosocial agents significantly outperform the baseline DQN, their performance levels exhibit a clear gap compared to our method and show slightly inferior stability in the later stages of learning. While empathic DQN surpasses the pure DQN baseline through its cognitive empathy mechanism, its effectiveness falls far short of models that directly act on rewards or optimization objectives. The proposed model achieves an average of 8.5 snowdrifts cleared after convergence stability, representing improvements of 2$\times$, 1.85$\times$, 1.37$\times$, and 1.27$\times$ compared to DQN (4.28), empathic DQN (4.59), inequity aversion (6.2), and prosocial agents (6.71), respectively. Furthermore, during later learning stages, the proposed model demonstrates more stable prosocial behavior, as evidenced by its final snowdrift-clearing standard deviation of merely 1.3964. This contrasts with the higher behavioral variability observed in baseline methods: empathic DQN ($\sigma$=1.5040), inequity aversion ($\sigma$=2.0199), and prosocial agents ($\sigma$=2.2685). In summary, the comparison with baseline methods in moral and empathy RL demonstrates the effectiveness of the proposed model in promoting the emergence of altruistic behaviors in social dilemma problems.

\subsubsection{Confusion Matrices under Different Empathy Levels}\label{con}

\begin{figure}[htbp]
    \centering
    \includegraphics[width=0.5\textwidth]{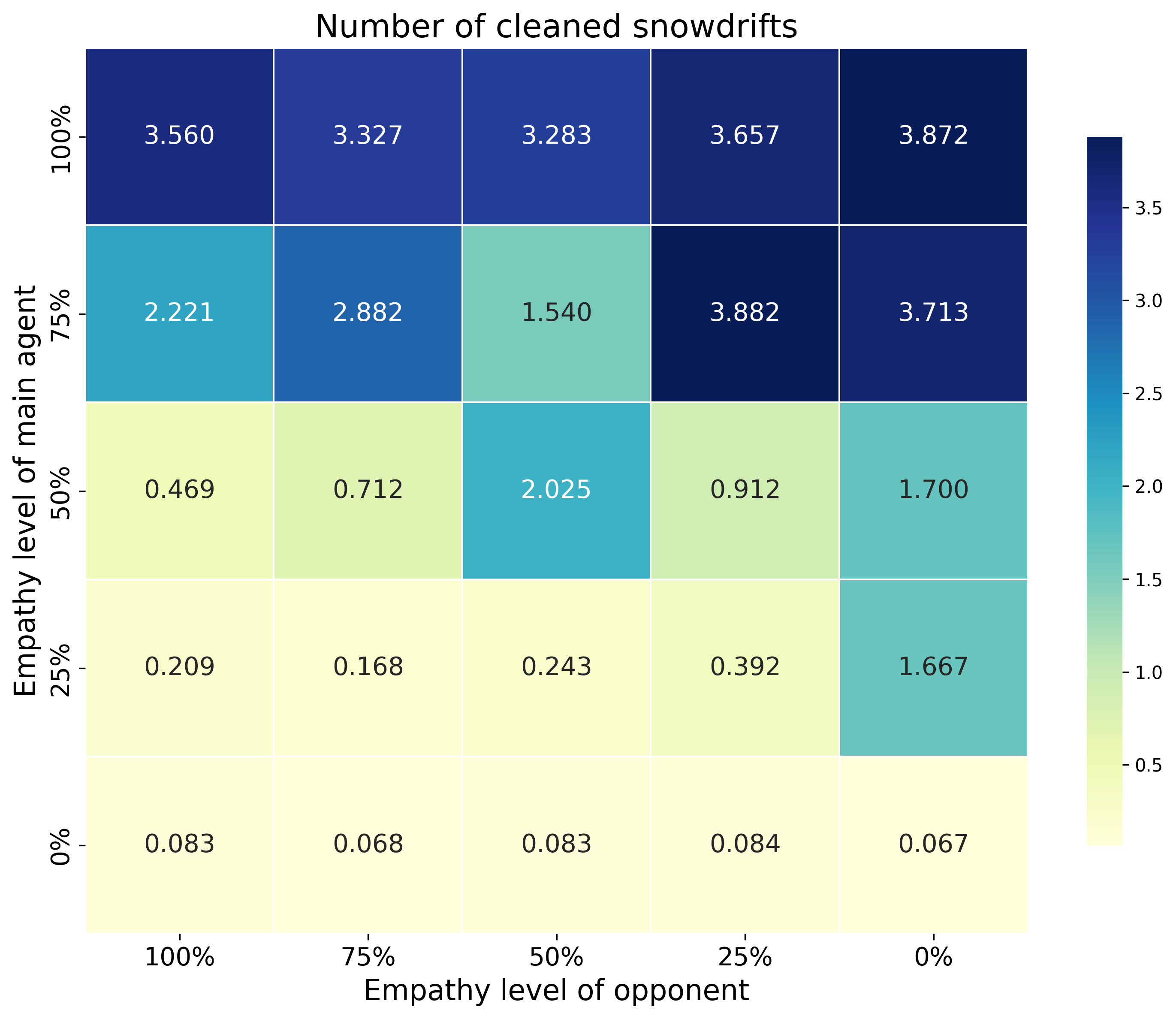}
    \caption{Snowdrift clearance by the main agent when gaming with differently empathic opponent.}
    \label{fig:14}
\end{figure}

To further analyze the impact of different empathy levels on agent behaviors, we assign different empathy levels to two agents in the snowdrift game scenario and record the number of snowdrifts cleared by the main agent during interactions, obtaining the confusion matrix shown in Fig.~\ref{fig:14}. We find that highly empathic  agents (the first row in Fig.~\ref{fig:14}) consistently exhibit prosocial altruistic behaviors by actively clearing snowdrifts, regardless of whether their counterparts cooperate. In contrast, agents lacking empathy (the last row in Fig.~\ref{fig:14}) almost never clear snowdrifts. Agents with intermediate empathy levels (25\%-75\%) demonstrate more complex and volatile behavioral patterns that depend on their interacting partners' empathy levels or behaviors.

In detail, for an agent with 75\% empathy, if its cooperator is with a higher level of empathy (100\%), the main agent tends to free-ride more frequently, thereby reducing its own inclination to perform clearing actions. When interacting with moderately empathic  cooperators (50\%), the 75\%-empathy agent employs a strategic delayed-clearing behavior: it waits until after the partner clears the snowdrift and displays negative emotional states, then performs clearing actions to alleviate the partner's distress. This conditional strategy results in reduced clearing frequency, as the agent's behavior becomes dependent on the partner's specific actions.
However, when cooperated with low-empathy agents (0\% or 25\%) who rarely clear snowdrifts, the 75\%-empathy agent resumes stable clearing behaviors, with frequency approaching that of 100\%-empathy agents. Overall, the confusion matrix reveals the dynamics of behavioral strategies during gaming interactions with different empathy levels, reflecting the complex game behavior and equilibrium phenomena.

\subsubsection{Partial Observability Experiments}

\begin{figure*}[htbp]
    \centering
    \includegraphics[width=1\textwidth]{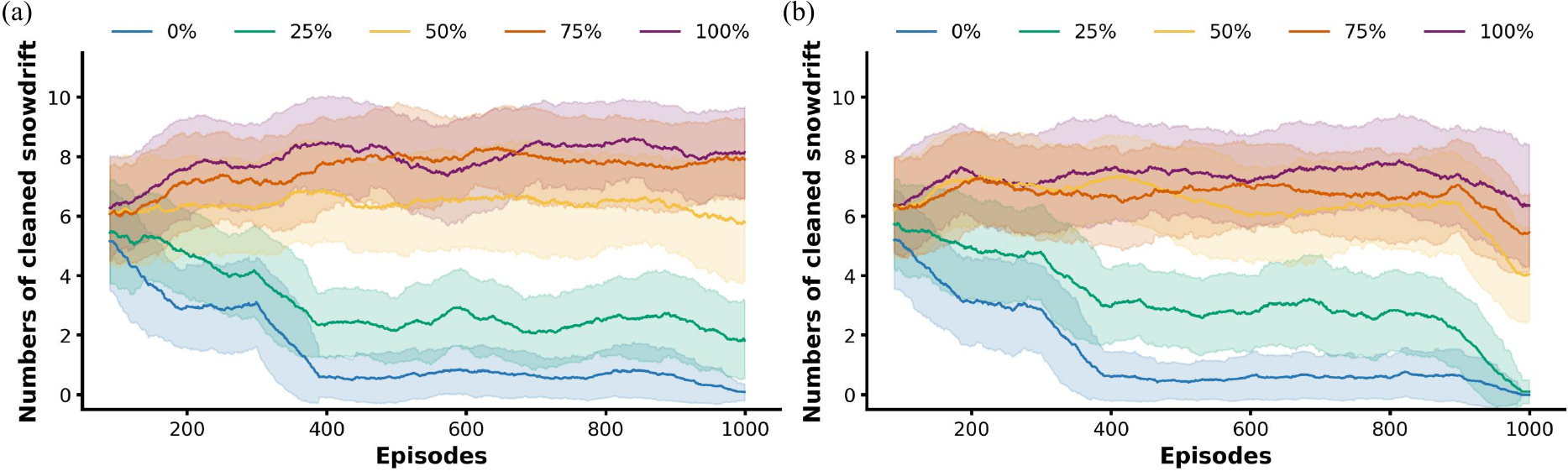}
    \caption{Experimental results under partial observation scenarios (a) 5*5 and (b) 3*3.}
    \label{15}
\end{figure*}

To evaluate the proposed model's performance under partial observability, we test the number of snowdrifts cleared (with different levels of empathy) under fully observed 8*8, and partially observed 5*5 and 3*3, as shown in Fig.~\ref{15} and Table~\ref{obs}. The results demonstrate a clear positive correlation between observational completeness and clearance efficiency. Notably, high-empathy agents (100\% and 75\%) exhibit remarkable generalization capability across observational conditions. Even with incomplete information, these agents maintained strong prosocial behavior by effectively utilizing local signals. However, in the most restricted 3×3 condition, clearance performance dropped sharply across all empathy levels, indicating that empathy-driven prosocial behavior requires minimum observational thresholds - when visibility becomes too limited to detect other agents, the advantages of empathy cannot be properly utilized. Interestingly, agents with intermediate empathy levels (50\% and 25\%) perform better in 5×5 than in 8×8 conditions. This aligns with our previous findings (see subsection~\ref{con}) that the agents may base their snow-clearing decisions on the behaviors of their interaction partners, while partial observability reduces their excessive reliance on others and surprisingly enhances their autonomous execution of prosocial behaviors.

\begin{table}[h]
\centering
\caption{Snowdrift Clearing Performance under Different Observation Conditions}
\label{obs}
\begin{tabular}{lccccc}
\toprule
 & 100\% & 75\% & 50\% & 25\% & 0\% \\
\midrule
Full observation & 8.53 & 8.417 & 4.597 & 1.271 & 0.293 \\
5×5 partial observation & 8.216 & 7.888 & 5.998 & 2.123 & 0.306 \\
3×3 partial observation & 6.793 & 5.993 & 4.911 & 0.946 & 0.233 \\
\bottomrule
\end{tabular}
\end{table}

\begin{figure}[htbp]
    \centering
    \includegraphics[width=0.45\textwidth]{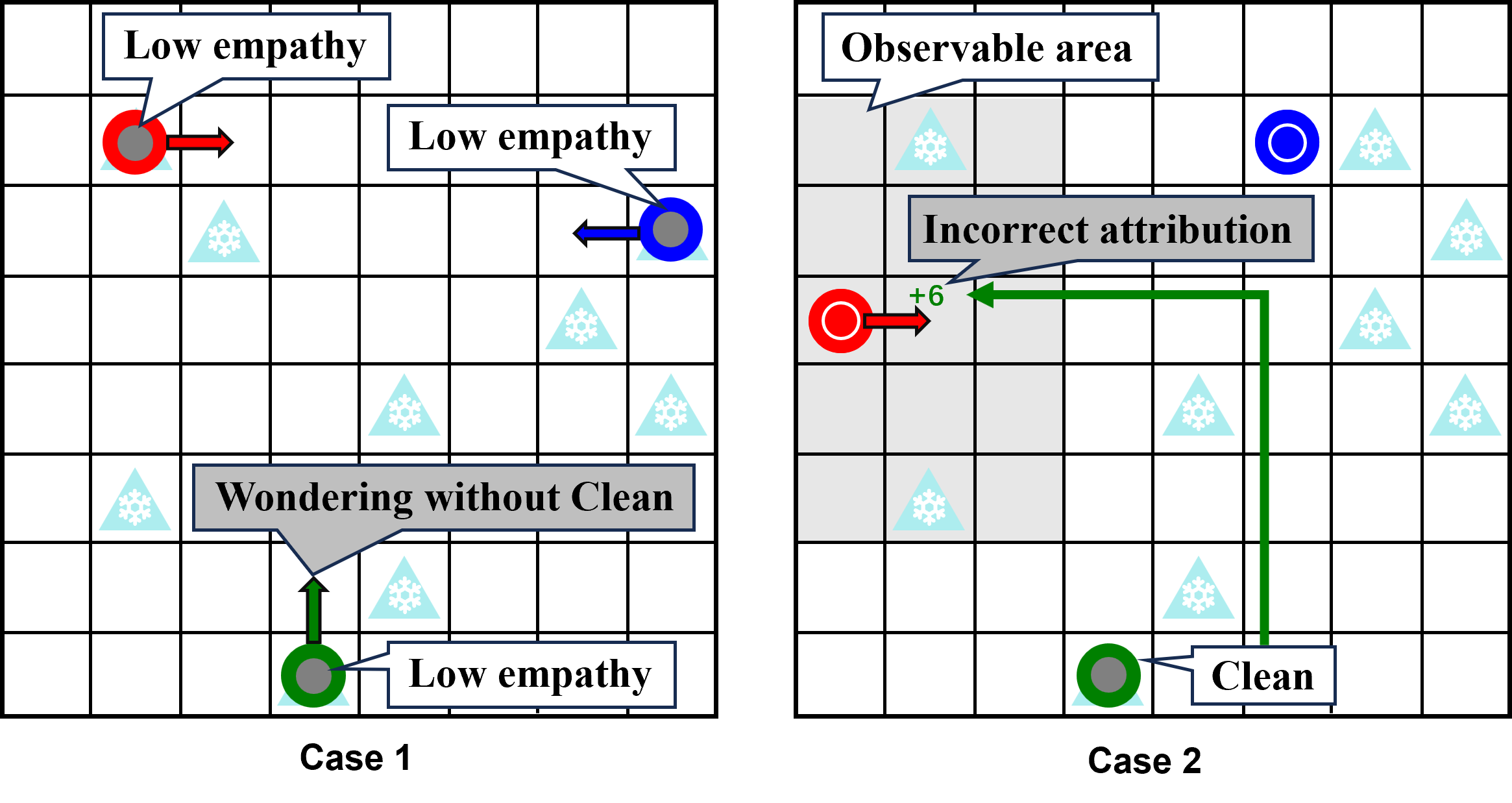}
    \caption{Instances of agent failure to perform prosocial behaviors (snowdrift remained uncleared).}
    \label{fig:16}
\end{figure}

In addition, through monitoring failure cases where agents failed to exhibit prosocial behaviors during multi-agent interactions, we show two representative edge cases (as shown in Fig.~\ref{fig:16}). In "Case 1", when multiple low-empathy agents simultaneously encountered snowdrifts, they mutually reinforced the expectation that others would clear the snowdrifts - enabling them to receive rewards while avoiding clearing penalties, ultimately creating a deadlock where all agents free-rode and none cleared. In "Case 2", under partial observability where agents couldn't perceive others' emotional states or strategies, increased strategic uncertainty led to local inaction rather than proactive clearing. These failures demonstrate that prosocial behavior requires both surpassing critical empathy thresholds and maintaining minimal observational capacity to assess others' situations.

\subsection{Application on Robot Self-other Resonance Experiments}

\begin{figure*}[htbp]
    \centering
    \includegraphics[width=1\textwidth]{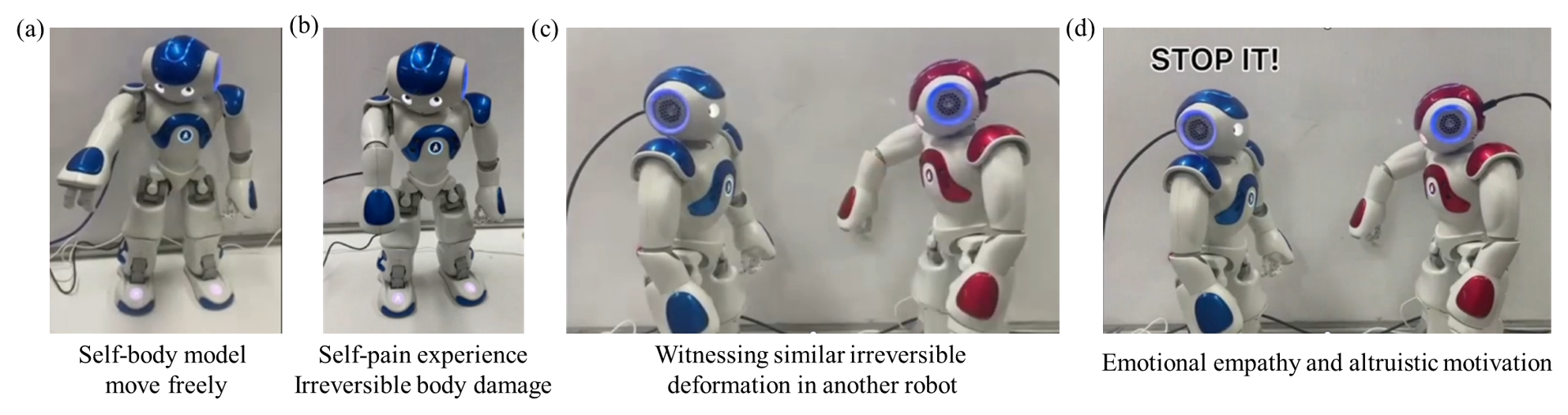}
    \caption{Application on robot self-other resonance experiment.}
    \label{fig:17}
\end{figure*}

The emotional empathy-driven altruistic decision-making model proposed in this paper holds potential for application in more humanized, emotionally social robots. However, real-world implementation faces additional challenges that require further integration with technologies such as vision- and language-based emotion recognition, biomimetic emotional expression, robot control, and human-robot interaction. This study preliminarily explores the application of the proposed model in enabling robots to empathize with other robots based on their own experiences. We utilize two Nao robots as experimental platforms, as shown in Fig.~\ref{fig:17}, where the blue robot possesses empathic capabilities. Referring to the robotic pain model~\cite{feng2022brain}, an injury is simulated by human bending of a robotic arm. The Nao robot establishes a self-body model to learn the association between pain emotion and movement.

First, the blue robot accumulates self-experience during a random exploration phase: in a normal bodily state, the mechanical arm could move freely, whereas forced bending resulted in an irreversible injured state, thereby generating a self-pain emotion (as depicted in Fig.~\ref{fig:17}(a) and (b)). Subsequently, when observing the red robot’s mechanical arm being bent, the same joint signals are input into the perception regions of the proposed model to achieve a first-person shared observational input (rather than vision-based state recognition). In this study, altruistic rescue behavior is simplified to emitting a distress call—voicing "Stop it!" The experimental process and results are illustrated in Fig.~\ref{fig:17}. When the blue robot observed the red robot’s arm bending, it trigger its own emotional experience and produce an "Stop it" vocalization.

\section{Conclusion}
\label{conc}
This paper presents an altruistic moral AI agent inspired by the emotional empathy mechanisms in the human brain, enabling the agent to empathize with others based on its own experiences and develop intrinsic motivation for altruism, particularly in moral dilemmas involving conflicts between self-interest and the interests of others. Specifically, we proposed a multi-brain area coordinated spiking neural network model that integrates the mirror neuron system for spontaneous empathy and regulates dopamine levels to drive altruistic decision making. Additionally, a moral reward system is designed based on moral 
deontology, combining intrinsic empathy-related dopamine levels with external self-task goals, facilitating consistent moral behavior that balances self-interest with altruism. In the designed moral decision-making experimental scenarios, emotional empathy spontaneously drives altruistic motivation, leading the agent to prioritize altruistic behavior even at the cost of sacrificing its own interests. The introduction of brain-inspired inhibitory neural populations allows for the regulation of different empathy levels, demonstrating that agents with higher empathy are more willing to sacrifice their interests to alleviate others' negative emotion, which aligns with psychological behavioral experiments.

This study provides a preliminary investigation into intrinsically altruistic behaviors driven by brain-inspired emotional empathy mechanism, currently focusing on externally observable emotional expressions and employing deontological ethics that prioritizes altruism to guide agents’ prosocial behaviors. The core scientific contribution lies in modeling the neurobiological mechanisms underlying empathy and moral decision making, ensuring both biological plausibility and effectiveness. However, real-world moral decision making exhibits far greater complexity—empathy may induce cognitive biases or be susceptible to manipulation, moral judgments inherently require multi-dimensional evaluation, and conflicts between moral norms frequently emerge. These critical challenges underscore the need for future research to systematically examine: the pluralistic nature of moral judgment (e.g., through the multidimensional framework of Moral Foundations Theory), dynamic interactions and value trade-offs between different moral dimensions, cross-cultural variations in empathy expression, and empathy-driven potential risks. Besides, we aim to explore empathy-enhanced robotic applications across complex domains including medical care, educational assistance, elderly companionship, service robotics, and collaborative robotics. We will progressively advance the development of ethically-aligned AI systems, ultimately establishing a safe and harmonious human-machine symbiotic ecosystem.

\ifCLASSOPTIONcompsoc
  \section*{Acknowledgments}
\else
  \section*{Acknowledgment}
\fi

This work is supported by the Strategic Priority Research Program of the Chinese Academy of Sciences (Grant No. XDB1010302), and the National Natural Science Foundation of China (Grant No. 62576341 and No. 32441109), and the Beijing Natural Science Foundation (Grant No.4252052), and the funding from Institute of Automation, Chinese Academy of Sciences (Grant No. E411230101), and the Beijing Major Science and Technology Project under Contract (Grant No. Z241100001324005).

\ifCLASSOPTIONcaptionsoff
  \newpage
\fi



%
\bibliographystyle{IEEEtran}
\bibliography{IEEEabrv,elife-sample}

%

\begin{IEEEbiography}[{\includegraphics[width=1in,height=1.25in,clip,keepaspectratio]{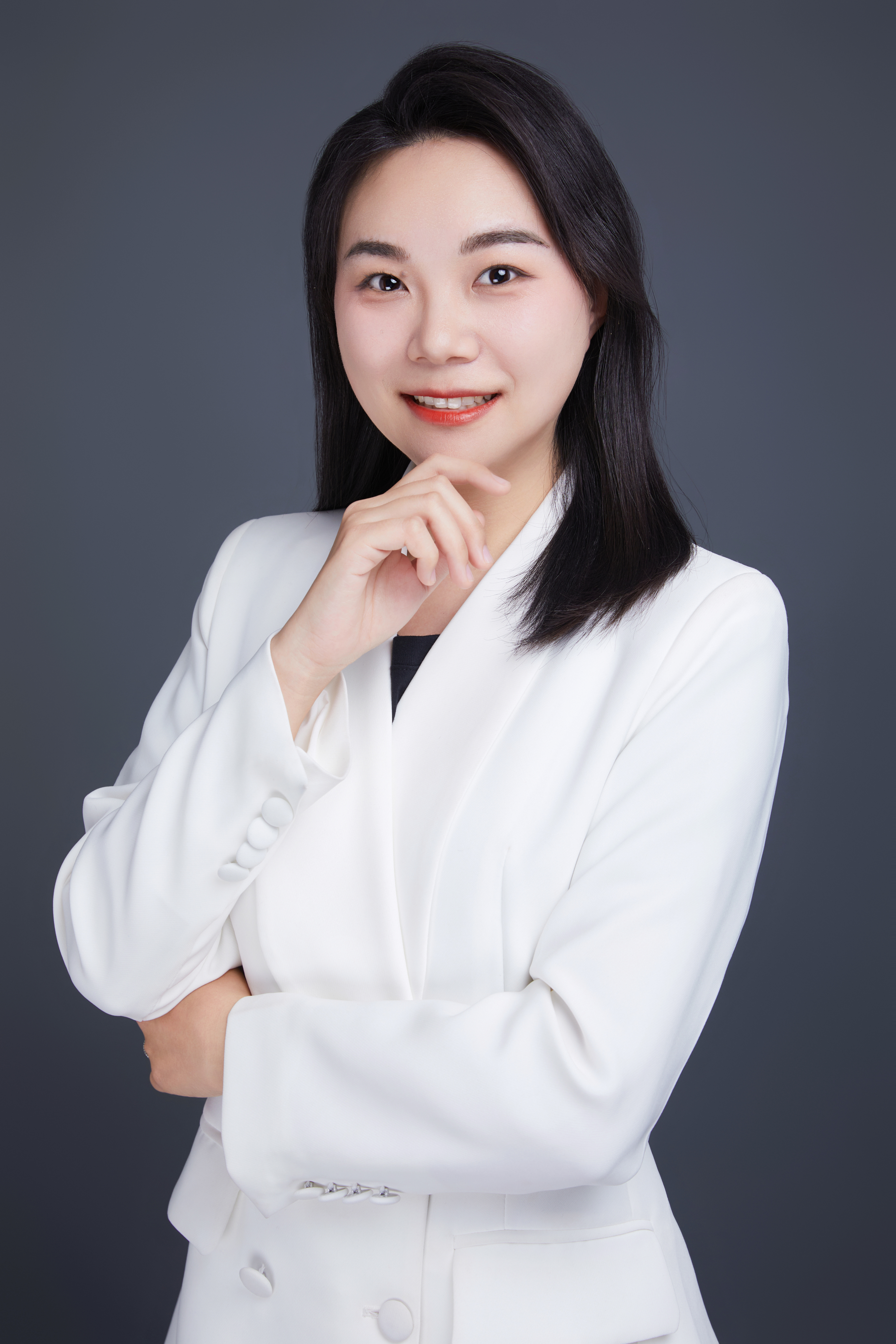}}]{Feifei Zhao}
is currently an Associate Professor in the Brain-inspired Cognitive AI Lab, Institute of Automation, Chinese Academy of Sciences (CASIA), China. She also serves as the research fellow of Beijing Key Laboratory of Safe AI and Superalignment, and Beijing Institute of AI Safety and Governance, and Long-term AI, China. Her current research interests include Brain-inspired Developmental and Evolutionary Spiking Neural Networks, AI Ethics and Safety.
\end{IEEEbiography}

\begin{IEEEbiography}[{\includegraphics[width=1in,height=1.25in,clip,keepaspectratio]{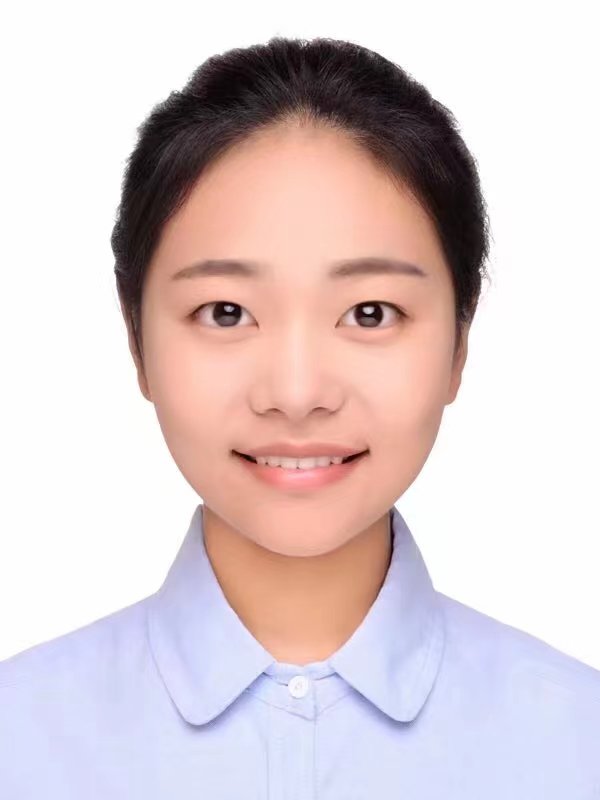}}]{Hui Feng} received the Ph.D. degree from the University of Chinese Academy of Sciences, Beijing, China, in 2024. Her research interests include brain-inspired spiking neural network models for emotional empathy and altruistic behavior.
\end{IEEEbiography}

\begin{IEEEbiography}[{\includegraphics[width=1in,height=1.25in,clip,keepaspectratio]{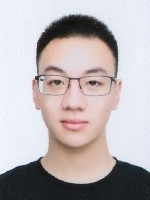}}]{Haibo Tong} is currently a Ph.D. Candidate in the Brain-inspired Cognitive AI Lab, Institute of Automation, Chinese Academy of Sciences (CASIA), China. His current research interests include spiking neural networks and AI safety.
\end{IEEEbiography}

\begin{IEEEbiography}[{\includegraphics[width=1in,height=1.25in,clip,keepaspectratio]{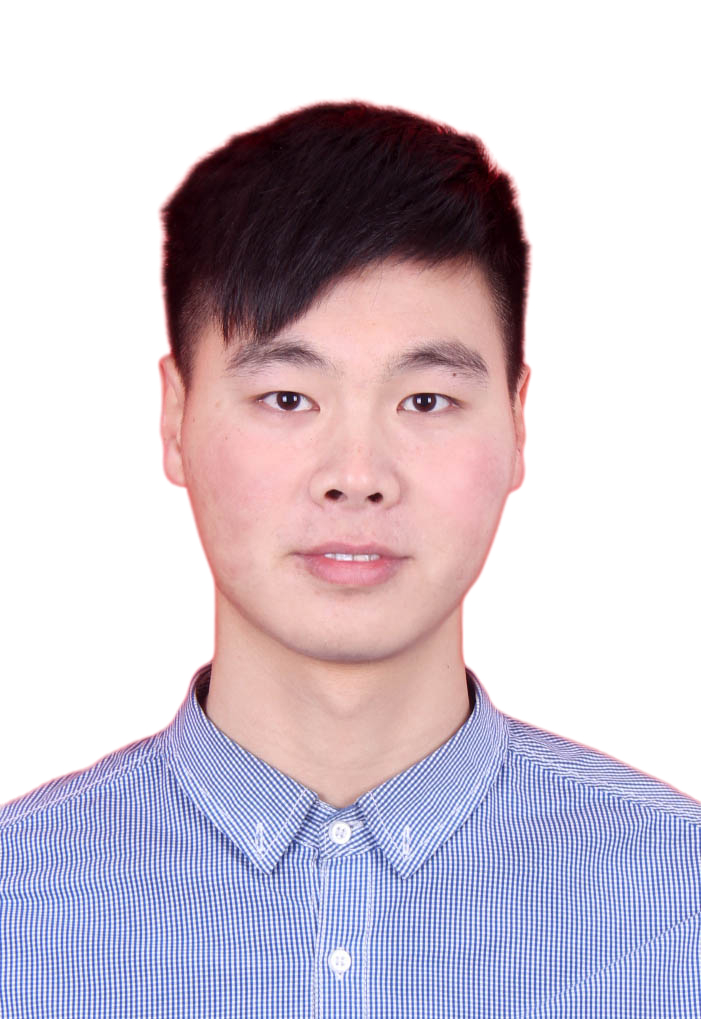}}]{Zhengqiang Han} is a Ph.D. Candidate of School of Humanities, University of Chinese Academy of Sciences, Beijing, China. He is also a student fellow in the International Research Center for AI Ethics and Governance. The Center is hosted at Institute of Automation, Chinese Academy of Sciences, Beijing, China. His current research interests include robot ethics and safety, and computational simulations of ethical principles.
\end{IEEEbiography}

\begin{IEEEbiography}[{\includegraphics[width=1in,height=1.25in,clip,keepaspectratio]{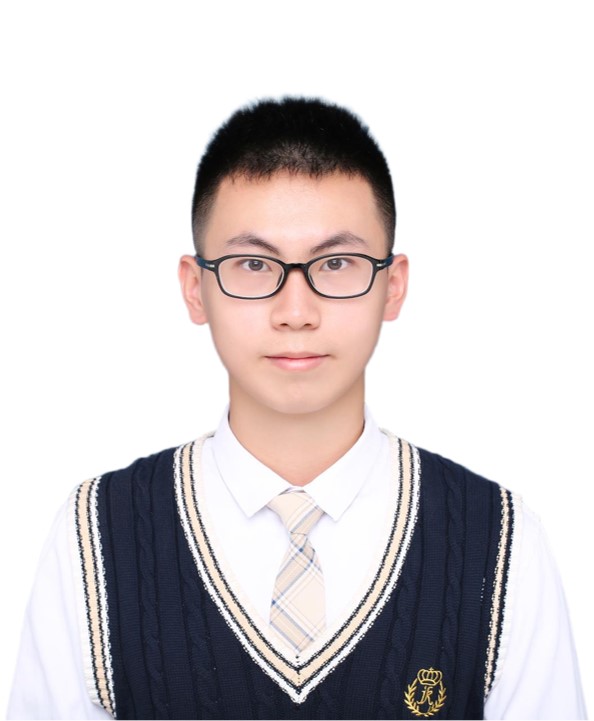}}]{Erliang Lin} is currently interning at Brain-inspired Cognitive AI Lab, Institute of Automation, Chinese Academic of Science (CASIA), China. His current research interests include brain-inspired artificial intelligence and AI value alignment.
\end{IEEEbiography}

\begin{IEEEbiography}[{\includegraphics[width=1in,height=1.25in,clip,keepaspectratio]{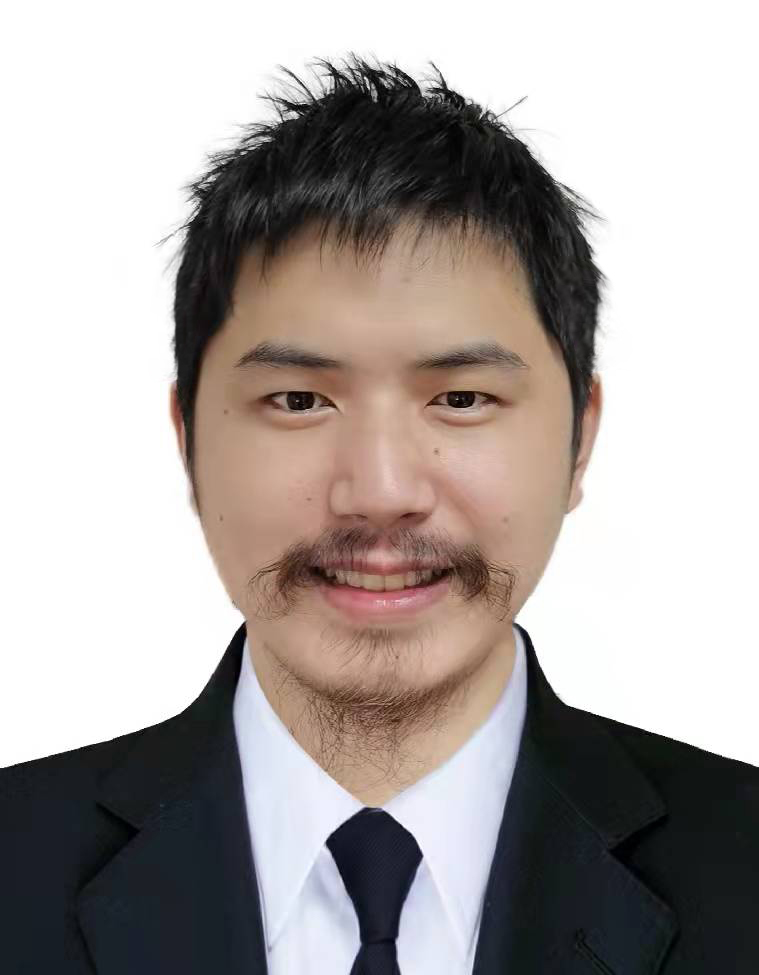}}]{Enmeng Lu} is currently a Senior Research Fellow at Beijing Institute of AI Safety and Governance (Beijing-AISI), and serves as the Director of Sustainable Development and AI Governance Research Center. He is also a Co-Director of Center for Long-term AI, China. His research focuses on the ethics, safety, and governance of AI, employing both technical and policy tools to address problems.

\end{IEEEbiography}

\begin{IEEEbiography}[{\includegraphics[width=1in,height=1.25in,clip,keepaspectratio]{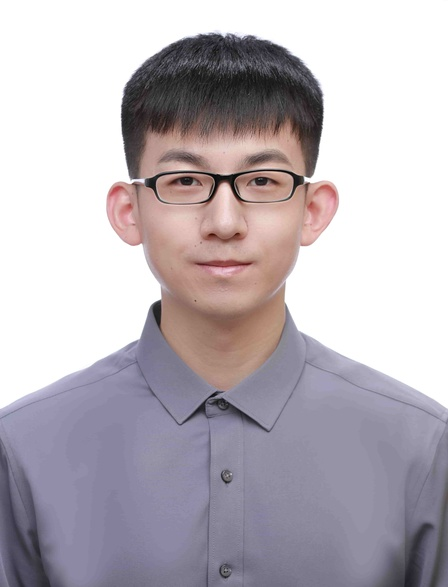}}]{Yinqian Sun} is currently an Assistant Professor in the Brain-inspired Cognitive AI Lab, Institute of Automation, Chinese Academy of Sciences (CASIA), China. He also serves as the research fellow of Beijing Key Laboratory of Safe AI and Superalignment, and Beijing Institute of AI Safety and Governance, and Long-term AI, China. His current research interests include Brain-inspired Decision-making Models, Brain-inspired Neural Robotics and Embodied AI.
\end{IEEEbiography}

\begin{IEEEbiography}[{\includegraphics[width=1in,height=1.25in,clip,keepaspectratio]{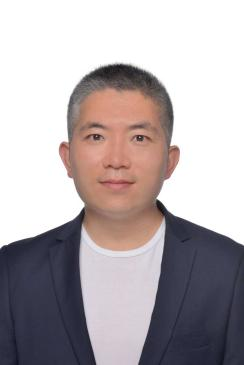}}]{Yi Zeng}
is a Professor and Director of the Brain-inspired Cognitive AI Lab at the Institute of Automation, Chinese Academy of Sciences (CASIA). He also serves as the Founding Director of the International Research Center for AI Ethics and Governance, and Director of the Beijing Key Laboratory of Safe AI and Superalignment. In addition, he is the Dean of the Beijing Institute of AI Safety and Governance.
He is also a Principal Investigator in the State Key Laboratory of Brain Cognition and Brain-inspired Intelligence Technology, Chinese Academy of Sciences, China, and a Professor in the Chinese Academy of Sciences, China. His research interests include Safety, Ethics, Governance of AI, 
Human-AI Super Co-alignment,
Brain and mind inspired Cognitive AI Models,
Brain-inspired Cognitive and Neural Robotics,
AI for Sustainable Development, 
as well as AI for International Peace and Security.
\end{IEEEbiography}




\end{document}